\documentclass[journal]{IEEEtran}
\usepackage{amsmath,amssymb,amsfonts}
\usepackage{cite}
\usepackage[ruled,linesnumbered]{algorithm2e}
\usepackage{graphicx}
\usepackage{caption}
\usepackage{lipsum}
\usepackage{amsthm}
\usepackage{color}
\usepackage{threeparttable}
\usepackage{cuted}
\usepackage{lipsum}
\usepackage{multicol}
\ifCLASSOPTIONcompsoc
  \usepackage[caption=false,font=normalsize,labelfont=sf,textfont=sf]{subfig}
\else
  \usepackage[caption=false,font=footnotesize]{subfig}
\fi

\usepackage{booktabs}
\begin{document}
\title{Centralized Cooperation for Connected and Automated Vehicles at Intersections by Proximal Policy Optimization}

\author{Yang Guan, Yangang Ren, Shengbo Eben Li*, Qi Sun, Laiquan Luo, and~Keqiang Li 
\thanks{Copyright (c) 2015 IEEE. Personal use of this material is permitted. However, permission to use this material for any other purposes must be obtained from the IEEE by sending a request to pubs-permissions@ieee.org.}
\thanks{Y. Guan, Y. Ren, S. E. Li, Q. Sun, L. Luo and K. Li, are with State Key Lab of Automotive Safety and Energy, School of Vehicle and Mobility, Tsinghua University, Beijing, 100084, China. They are also with Center for Intelligent Connected Vehicles and Transportation, Tsinghua University, Beijing, China. {\tt\small Email: (guany17, ryg18)@mails.tsinghua.edu.cn; qisun@mail.tsinghua.edu.cn;
(lisb04, luolaiquan)@gmail.com; likq@tsinghua.edu.cn}.
}
\thanks{*The corresponding author is Shengbo Eben Li. All questions about this paper should be sent to email {\tt\small lisb04@gmail.com}.}}

\maketitle
\pagestyle{empty}
\thispagestyle{empty}

\begin{abstract}
Connected vehicles will change the modes of future transportation management and organization, especially at an intersection without traffic light. Centralized coordination methods globally coordinate vehicles approaching the intersection from all sections by considering their states altogether. However, they need substantial computation resources since they own a centralized controller to optimize the trajectories for all approaching vehicles in real-time.
In this paper, we propose a centralized coordination scheme of automated vehicles at an intersection without traffic signals using reinforcement learning (RL) to address low computation efficiency suffered by current centralized coordination methods.
We first propose an RL training algorithm, model accelerated proximal policy optimization (MA-PPO), which incorporates a prior model into proximal policy optimization (PPO) algorithm to accelerate the learning process in terms of sample efficiency.
Then we present the design of state, action and reward to formulate centralized coordination as an RL problem.
Finally, we train a coordinate policy in a simulation setting and compare computing time and traffic efficiency with a coordination scheme based on model predictive control (MPC) method. Results show that our method spends only 1/400 of the computing time of MPC and increase the efficiency of the intersection by 4.5 times.
\end{abstract}


\IEEEpeerreviewmaketitle

\section{Introduction}
\IEEEPARstart{T}{he} increasing demand for mobility poses great challenges to road transport. The connected and automated vehicles are attracting extensive attention, due to its potential to benefit traffic safety, efficiency and economy \cite{li2015eco, tang2017novel, liu2019adaptive}.
The widely studied, but also simplified, version of connected vehicle cooperation is the platoon control system on the highway. Platoon control aims to ensure that a group of connected vehicles in the same lane move at a harmonized longitudinal speed while maintaining desired inter-vehicle spaces\cite{wu2019distributed, gao2018distributed, li2017robustness, guo2020distributed}.
As a typical scenario in urban areas, the intersection is more complex and challenging for multi-vehicle coordination than that on the highway. At the intersection, vehicles enter from different intersection entrances, cross their specific trajectories at the intersection zone, and leave the intersection at different exits. The complex conflict relationship between vehicles results in complicated vehicle decisions to avoid collisions at the intersection, which needs complicated design to guarantee traffic safety while improving traffic efficiency. Traffic signal control is commonly considered as the most effective method to resolve multi-vehicle coordination at intersections, and various strategies for urban traffic management have been developed. Goodall et al. developed a decentralized fully adaptive traffic control algorithm to optimize traffic signal timing \cite{goodall2013traffic}. Feng et al. presented a real-time adaptive signal phase allocation algorithm using connected vehicle data, which optimizes the phase sequence and duration by solving a two-level optimization problem \cite{feng2015real}. 
These signal control strategies can only partially improve the traffic flow if all approaches to the intersection are not equally congested, and they cannot eliminate the stop delay of vehicles at intersections.
Recently, several studies have started to focus on methods without using traffic signals for intersection coordination.

Currently, most of the existing studies without the traffic signal focused on centralized coordination methods which utilize the global information of the whole intersection to centralizedly organize the motion of all approaching vehicles.
Lee and Park proposed a system that enables cooperation between automated vehicles and infrastructures for effective intersection operations \cite{lee2012development}. The system tries to avoid the presence of any pair of conflicting vehicles in the intersection area at the same time by minimizing potential overlaps of trajectories of vehicles coming from all conflicting approaches at the intersection, then seeks a safe maneuver for each vehicle approaching the intersection and manipulates each of them. To construct the objective function of the nonlinear constrained optimization problem, they predicted the trajectory of each vehicle by fixing an acceleration from its current position to the end of the intersection. In their optimization problem, the computation requirements usually increase rapidly with the growth of the number of vehicles and should be determined carefully to ensure that the feasibility of the real-time implementation is guaranteed.

Kamal et al. proposed a vehicle-intersection coordination scheme (VICS) without using any traffic light \cite{kamal2014vehicle}. The scheme efficiently utilizes the intersection area by preventing each pair of conflicting vehicles from approaching their cross-collision point (CCP) at the same time, instead of reserving the whole intersection area for the conflicting vehicles successively. In this scheme, a risk function is proposed to quantify the risk of a collision of a pair of vehicles around their CCP. If two conflicting vehicles are very close to their CCP, the risk function returns a high value, and if at least one vehicle is far from the CCP, it returns a negligible value. Considering states of all vehicles, they solved a constrained nonlinear optimization problem under a model predictive control (MPC) framework in order to let the vehicles cross the intersection rapidly by minimizing the total quantified risks of all vehicle pairs. However, MPC is usually computationally demanding. In their experiment, the average computation time is found to be about 1.76s per iteration, which is obviously not practical in real world application.

To reduce computational overhead, Dai et al. first quantitatively analyzed the characteristics of vehicle movement and transformed their factors into a convex objective function. Furthermore, they defined the decision zone and divided an intersection into multiple collision areas. Then they designed a schedule rule to determine the priority of the vehicles in each collision area, which linearizes the collision constraints. On this basis, they transformed the traffic control model into a convex optimization, which can be solved efficiently in real time \cite{dai2016quality}. However, an additional algorithm, in the form of priority rules for each collision area, is required to deal with the linearization of nonlinear constraints.

Reinforcement learning (RL) has the potential to solve low computation efficiency suffered by current centralized coordination methods. Given a reward function, RL learns a policy by trial-and-error within a simulated environment or the real world to maximize the sum of future rewards. Then the learned policy can be used to get real time decisions compared with solving an optimization problem in each step. Existing RL research on autonomous driving mostly focus on the intelligence of single-vehicle driving in relatively simple traffic scenarios \cite{wolf2017learning, ruiming2018end,jaritz2018end, dosovitskiy2017carla,duan2019hierarchical,kendall2019learning, xiong2016combining}. In this paper, we employ RL as our method for centralized coordination of multiple connected vehicles to realize autonomous passing at intersections without traffic signals. We first propose an RL training algorithm, model accelerated proximal policy optimization (MA-PPO), which incorporates a prior model into proximal policy optimization (PPO) algorithm to enhance sample efficiency. Then we present the design of state, action and rewards to formulate centralized coordination as an RL problem, where the rewards include a collision punishment and a step punishment for safety and efficiency considerations. Finally, a simulation-based case study implemented on a four-way single-lane approach intersection showed that our method spends only 1/400 of the computing time of VICS and increases the efficiency of the intersection by 4.5 times.

The main contributions of this paper are summarized as follows. First, we use reinforcement learning in centralized coordination of connected vehicles at an intersection without traffic signals and reduce computation overhead significantly. Second, we propose model accelerated proximal policy optimization algorithm, which incorporates a prior model into proximal policy optimization to accelerate the learning process in terms of sample efficiency. Third, we conduct extensive simulations, and the results demonstrate the superiority of our proposed method.

The rest of this paper is organized as follows. Section \ref{Preliminaries} introduces the preliminaries of Markov decision process (MDP) and policy gradient methods. Section \ref{Model accelerated PPO} proposes MA-PPO, which is an improvement based on PPO and model-based RL methods. Section \ref{problem statement} illustrates our problem statement and methodology and section \ref{Experiment settings} looks into experimental settings and illustrates results. Last section \ref{Conclusion} summaries this work.

\section{Preliminaries of RL}\label{Preliminaries}
Consider an infinite-horizon discounted MDP, defined by the tuple ($\mathcal{S}, \mathcal{A}, p, r, d^0, \gamma$), where $\mathcal{S}$ and $\mathcal{A}$ are state and action spaces respectively. We suppose that $\mathcal{S}$ and $\mathcal{A}$ are compact subsets of $\mathbb{R}^n$ and $\mathbb{R}^m$. $p: \mathcal{S} \times \mathcal{A} \times \mathcal{S} \rightarrow \mathbb{R}$ is the transition probability distribution. $r: \mathcal{S} \times \mathcal{A} \times \mathcal{S} \rightarrow \mathbb{R}$ is the reward function, $d^0: \mathcal{S} \rightarrow \mathbb{R}$ is the distribution of the initial state $s_0$, and $\gamma \in (0, 1)$ is the discount factor.

Let $\pi$ denote a stochastic policy $\pi: \mathcal{S}\times \mathcal{A} \rightarrow [0, 1]$, we seek to learn the optimal policy ${\pi}^*$ which has maximum values $v^{\pi^*}(s)$ for all $s \in \mathcal{S}$, where the value function $v^{\pi}(s)$ is the expected sum of discounted rewards from a state when following policy $\pi$:
\begin{equation}
\nonumber
    v^{\pi}(s) =\mathbb{E}_{a_{t}, s_{t+1}, \ldots}\left\{\sum_{l=t}^{\infty} \gamma^{l-t}r_l  | s_{t}=s\right\},
\end{equation}
where $a_{t} \sim \pi\left(a_{t} | s_{t}\right), s_{t+1} \sim p\left(s_{t+1} | s_{t}, a_{t}\right)$ and $r_t:= r\left(s_{t}, a_{t}, s_{t+1}\right)$ for short. Similarly, we use the following standard definition of the state-action value function $q^{\pi}(s,a)$:
\begin{equation}
\nonumber
    q^{\pi}(s,a) =\mathbb{E}_{s_{t+1}, a_{t+1}, \ldots}\left\{\sum_{l=t}^{\infty} \gamma^{l-t} r_l | s_{t}=s, a_{t}=a\right\}.
\end{equation}

\subsection{Vanilla policy gradient} 
In practice, finding the optimal actions for all states is impractical for problems with large state space. We instead use a parameterized policy $\pi_{\theta}(a|s)$ and try to seek the optimal parameter vector $\theta$. For the same reason, state-value function is parameterizd as $V(s, w)$ with a parameter vector $w$.
Policy optimization methods seek to find optimal $\theta^*$ which maximize average performance of policy $\pi_{\theta}$, i.e.,
\begin{equation}\label{objectivefunction}
    J(\theta) =\mathbb{E}_{s_{0}, a_{0}, \ldots}\left\{\sum_{t=0}^{\infty} \gamma^{t}r_t \right\},
\end{equation}
where $s_{0} \sim d^0(s_0), a_t \sim \pi_{\theta}(a_t|s_t), s_{t+1} \sim p\left(s_{t+1} | s_{t}, a_{t}\right)$ and $r_t:= r\left(s_{t}, a_{t}, s_{t+1}\right)$.
    
Vanilla methods optimize \eqref{objectivefunction} by stochastic policy gradient \cite{sutton2000policy}. Its gradient is shown as
\begin{equation}\label{vanilla}
    \nabla_{\theta} J(\theta) =\mathbb{E}_{s \sim d^{\gamma} , a \sim \pi_{\theta}}\left\{\nabla_{\theta} \log \pi_{\theta}(a | s) q^{\pi_{\theta}}(s, a)\right\},
\end{equation}
where
$d^{\gamma}$ is called discounted visiting frequency, which in practice is usually replaced with the stationary state distribution under $\pi_{\theta}$ denoted by $d^{\pi_{\theta}}$  \cite{thomas2014bias}. Combined with the baseline technique \cite{sutton2018reinforcement}, we can write \eqref{vanilla} in format
\begin{equation}
\nonumber
    \begin{aligned}
    \nabla_{\theta} J(\theta) \approx  \mathbb{E}_{s \sim d^{\pi_{\theta}} , a \sim \pi_{\theta}}\left\{\nabla_{\theta} \log \pi_{\theta}(a | s) A^{\pi_{\theta}}(s, a)\right\},
    \end{aligned}
\end{equation}
where $A^{\pi_{\theta}}(s, a) := q^{\pi_{\theta}}(s, a) - v^{\pi_{\theta}}(s)$ is the advantage function which could be estimated by several methods \cite{schulman2015high}.

\subsection{Trust region method}
While the vanilla policy gradient is simple to implement, it often leads to destructively large policy updates. Trust region policy optimization (TRPO) optimizes a lower bound of \eqref{objectivefunction} to guarantee performance improvement, i.e.,
\begin{equation}\label{origTRPO}
\underset{\theta}{\operatorname{max}} \mathbb{E}_{s,a}\left[\frac{\pi_{\theta}\left(a | s\right)}{\pi_{\theta_{\text {old}}}\left(a | s\right)} A^{\pi_{\theta_{\text{old}}}}-\beta \mathrm{KL}\left[\pi_{\theta_{\text {old}}}\left(\cdot | s\right), \pi_{\theta}\left(\cdot | s\right)\right]\right].
\end{equation}

However, it is difficult to choose a single value of $\beta$ that performs well across different problems, TRPO uses a constraint instead, shown as
\begin{equation}
\nonumber
\begin{aligned}
  &\underset{\theta}{\operatorname{max}} \quad \mathbb{E}_{s,a}\left[\frac{\pi_{\theta}\left(a | s\right)}{\pi_{\theta_{\text {old}}}\left(a | s\right)} A^{\pi_{\theta_{\text{old}}}}\right]\\
&\operatorname{s.t.} \quad \mathbb{E}_{s}\left[D_{\mathrm{KL}}\left(\pi_{\theta_{\text {old}}}(\cdot | s) \| \pi_{\theta}(\cdot | s)\right)\right] \leq \delta,
\end{aligned}
\end{equation}
where $\delta$ is a bound on the KL divergence between the new policy and the old policy, $\mathbb{E}_{s,a}\left[\ldots\right]:=\mathbb{E}_{s \sim d^{\pi_{\theta_{\text{old}}}}, a \sim \pi_{\theta_{\text{old}}}}\left[\ldots\right]$,  $\mathbb{E}_{s}\left[\ldots\right]:=\mathbb{E}_{s \sim d^{\pi_{\theta_{\text{old}}}}}\left[\ldots\right]$ and $A^{\pi_{\theta_{\text{old}}}} := A^{\pi_{\theta_{\text{old}}}}(s,a)$.
To solve the constrained optimization problem, TRPO linearizes the objective function and transforms the constraint into a quadratic constraint. Because it is time-consuming to solve the inverse of Hessian matrix for large-scale problems, in each iteration, the authors use the conjugate gradient method to solve a system of linear equations to obtain the feasible ascent direction, and then obtain the optimal solution through the line search method.
TRPO can be regarded as natural policy gradient methods \cite{kakade2002natural}. It finds the steepest policy gradient in the fisher matrix normed space rather than the euclidean space, which helps to reduce impact of policy parameterization and stabilize the learning process.
However, the conjugate gradient method leads to a fixed number of inner iterations every time the feasible direction is solved, which increases the complexity of the algorithm.

\section{Model accelerated PPO}\label{Model accelerated PPO}
\subsection{Proximal policy optimization}
In this paper, the PPO algorithm is employed as our baseline. It is inspired by TRPO and has two main differences, namely, unconstrained surrogate objective function and generalized advantage estimation.
\subsubsection{Unconstrained surrogate objective function}
Observing monotonic policy improvement requires punishment of policy deviation from the theroy of TRPO, PPO alternatively constructs an unconstrained surrogate objective function to remove the incentive for large policy updates. Its objective is shown as
\begin{equation}\label{ppoobjective}
J^{\text{ppo}}(\theta)=\mathbb{E}_{s,a}\left[\min \left(\rho_{\theta} A^{\pi_{\theta_{\text{old}}}}, \operatorname{clip}\left(\rho_{\theta}, 1-\epsilon, 1+\epsilon\right) A^{\pi_{\theta_{\text{old}}}}\right)\right],
\end{equation}
where $\rho_{\theta} = \frac{\pi_{\theta}\left(a | s\right)}{\pi_{\theta_{\text {old}}}\left(a | s\right)}$. When $A^{\pi_{\theta_{\text{old}}}} > 0$, the term $\rho_{\theta}$ would tend to be much larger than 1 to make the performance as high as possible, which leads to unstable learning. The objective of PPO \eqref{ppoobjective} cuts this motivation by clipping $\rho$ within $1 + \epsilon$. Same situation is with $A^{\pi_{\theta_{\text{old}}}} < 0$.

\subsubsection{Generalized advantage estimation}
The advantage function is necessary for the calculation of the PPO policy gradient, and it can be estimated by
\begin{equation}
\nonumber
    \hat{A}^{\pi}(s, a) = \hat{Q}^{\pi}(s, a) - V(s, w),
\end{equation}
where $\hat{Q}^{\pi}(s, a)$ is the action-value function estimated by samples, $V(s, w)$ is the approximation of the state-value function. TRPO uses Monte Carlo methods to construct $\hat{Q}^{\pi}(s, a)$, i.e.,
\begin{equation}
\nonumber
    \hat{Q}^{\pi}(s_t, a_t) = \sum_{l=t}^{\infty}\gamma^{l-t}r_l.
\end{equation}
It is unbiased but suffers from high variance. Actor-critic methods use one-step bootstrapping to form $\hat{Q}^{\pi}(s, a)$, i.e.,
\begin{equation}
\nonumber
    \hat{Q}^{\pi}(s_t, a_t) = r_t + \gamma V(s_{t+1}, w),
\end{equation}
which is biased but has low variance. 

Generalized advantage estimation uses the linear combination of $n$-step bootstrapping to obtain both low bias and low variance, which is shown as
\begin{equation}
\nonumber
     \hat{Q}^{\pi}(s_t, a_t) = \sum_{l=t}^{\infty}(\gamma\lambda)^{l-t}\delta_l + V(s_t, w),
\end{equation}
where $\delta_l$ is the TD error,
\begin{equation}
\nonumber
  \delta_l = r_l + \gamma V(s_{l+1}, w) - V(s_{l}, w).
\end{equation}

Compared with TRPO, PPO is much simpler and faster to implement because it is a first-order optimization algorithm and has better convergence speed when it is combined with generalized advantage estimation. However, PPO is an on-policy method and inevitably has high sample complexity.

\subsection{Model-based RL}\label{modelsection}
To reduce sample complexity, recent RL algorithms have been proposed to incorporate a given or learned environment model as an additional source of information, where the environment models are models of the true transition dynamics $p$. Generally, there are two ways to use environment models: value gradient methods, or using models for imagination.

Value gradient methods link together the policy, model, and reward function to compute an analytic policy gradient by backpropagation of reward along a trajectory \cite{deisenroth2011pilco, grondman2015online, heess2015learning}. A major limitation of this approach is that the dynamic model can only be used to retrieve information already presented in observed data, and albeit with lower variance, the actual improvement in efficiency is relatively small. Alternatively, the given or learned model can also be used for imagination. This usage can be naturally incorporated in the model-free RL framework. However, the inconsistency between models and true dynamics can lead to large errors when the model-generated rollouts are too long \cite{kurutach2018model, feinberg2018model}.

\subsection{Model accelerated PPO}
PPO is a model-free on-policy RL algorithm. Model-free means it knows nothing about the environment and can only learn from interactions with the environment. As a result, it inevitably requires a large amount of experience data to learn the value function and policy from scratch. 
Even worse, the on-policy property makes experiences produced by previously trained policies useless, which aggravates sample inefficiency. The large demand for on-policy samples will lead to a large amount of time to interact with the environment, as well as the safety issues of training in the real world.
This is our motivation to accelerate PPO in terms of sample efficiency.

In order to alleviate the low sample efficiency suffered by PPO, we employ the environment model in the training process, and propose model accelerated PPO (MA-PPO), which can not only learn from the interaction with the environment, but also learn from the model. In MA-PPO, the environment model is used for imagination, that is, we use the model to generate virtual samples, and then use the algorithm to learn from these samples. In this way, the objective function of PPO can be used without any change, and the environment model and PPO algorithm can be naturally combined. By learning from the data generated by the environment model, MA-PPO can reduce the dependence on real samples generated by interactions with a simulator or the real world. When the model error is small, MA-PPO can achieve better performance using fewer real samples, enhancing sample efficiency of PPO.

A big concern is that although learning from the data generated by the model can greatly improve the sample efficiency, the final performance of the algorithm will be affected by the existence of model errors. Considering this problem, MA-PPO controls for uncertainty in the model by only allowing imaging to a fixed depth, because short rollouts are more likely to reflect the real dynamics, reducing the opportunities for policies to rely on inaccuracies of model predictions. The depth is set to be a hyperparameter and can be determined according to the error of the model in problems.

MA-PPO algorithm is shown in algorithm \ref{alg.modelaccppo}. The first line of the algorithm is used to initialize network parameters and algorithm parameters. The second line starts the main loop of the algorithm, which is divided into two main parts. The first part describes the algorithm learning from samples generated by interacting with the environment (lines 3-16). The emphasis is on estimating the advantage function and the value target (lines 7-8) using generalized advantage estimation method, and then they are used to calculate the gradient of the policy and the value function respectively (lines 11-16). The second part describes the algorithm learning through virtual samples generated by the environment model (lines 17-28). First, a batch of data is extracted from the buffer (line 17) , and then the model and policy are used to extend each data point in the batch to the depth $D$, where the depth is used to control uncertainties in the model. The estimated advantage functions and value function targets are calculated through these virtual samples (lines 18-23). Finally, the policy gradient and value function gradient are calculated, and the corresponding parameters are updated (lines 25-28).

\begin{algorithm}[tb]
   \caption{MA-PPO}
   \label{alg.modelaccppo}
   Randomly initialize value network $V(s, w)$ and policy network $\pi_{\theta}$ with weights $w$ and $\theta$, set $\lambda, \gamma, \epsilon$, total timesteps $T_{total}$, batch size $B$, minibatch size $MB$, epoch $U$, model rollout depth $D$, buffer $\mathcal{B}=\varnothing$
   
   \For{iteration = 1, 2, $\cdots$, $T_{total}/B$}{
       Run several episodes using policy $\pi_{\theta}$ to collect $B$ timesteps $\mathcal{D}=(s_i, a_i, r_i, s_{i+1})_{i=0:B-1}$
       
       $\mathcal{B}=\mathcal{B} \cup \mathcal{D}$
       
       Calculate TD errors $\delta_i$, $i=0,1,\cdots,B-1$
       
       \For{i=0, 1, $\cdots$, B-1}{
           $\hat{A}_i = \sum_{l \ge i}^{End\_episode}(\lambda\gamma)^{l-i}\delta_l$
       
           Estimate value target $\hat{V}_i = \hat{A}_i + V(s_i, w)$
       }
       
       $\pi_{\theta_{\text{old}}} \gets \pi_{\theta}$
       
       \For{epoch = 1, 2, $\cdots$, $U$}{
           $J^{\text{PPO}}(\theta) = \frac{1}{B}\sum_{i=1}^{B}\min (\frac{\pi_{\theta}(a_i|s_i)}{\pi_{\theta_{\text{old}}}(a_i|s_i)}\hat{A}_i, \text{clip}(\frac{\pi_{\theta}(a_i|s_i)}{\pi_{\theta_{\text{old}}}(a_i|s_i)},1-\epsilon, 1+\epsilon)\hat{A}_i)$
           
           Update $\theta$ by $\nabla_{\theta}J^{\text{PPO}}$
           
           $J^{\text{Value}}(w) = -\frac{1}{B}\sum_{i=1}^{B}(\hat{V}_i - V(s_i, w))^2$
           
           Update $w$ by $\nabla_{w}J^{\text{Value}}$
      }    
      
      Sample $\mathcal{D}=(s_i, a_i, r_i, s_{i+1})_{i=0:B-1}$ from $\mathcal{B}$
      
      \For{i=0, 1, $\cdots$, B-1}{
           Extend $\hat{s}_0=s_i$ by the policy $\pi_{\theta}$ and the environment model to the depth $D$.
           Denote virtual samples as $(\hat{s}_t, \hat{a}_t, \hat{r}_t, \hat{s}_{t+1})_{t=0: D-1}$
           
           Calculate TD errors $\hat{\delta}_t, t=0,1,\cdots,D-1$ by virtual samples
           
           $\hat{A}_i = \sum_{l=0}^{D}(\lambda\gamma)^{l}\hat{\delta}_l$
       
           Estimate value target $\hat{V}_i = \hat{A}_i + V(s_i, w)$
       }
       
      $\pi_{\theta_{\text{old}}} \gets \pi_{\theta}$
       
      \For{epoch = 1, 2, $\cdots$, $U$}{
          Update $\theta$ by $\nabla_{\theta}J^{\text{PPO}}$
           
          Update $w$ by $\nabla_{w}J^{\text{Value}}$
       }
    }
\end{algorithm}

\section{Problem statement and formulation}\label{problem statement}
\subsection{Problem statement}
\begin{figure}[htbp]
\centerline{\includegraphics[width=0.8\linewidth]{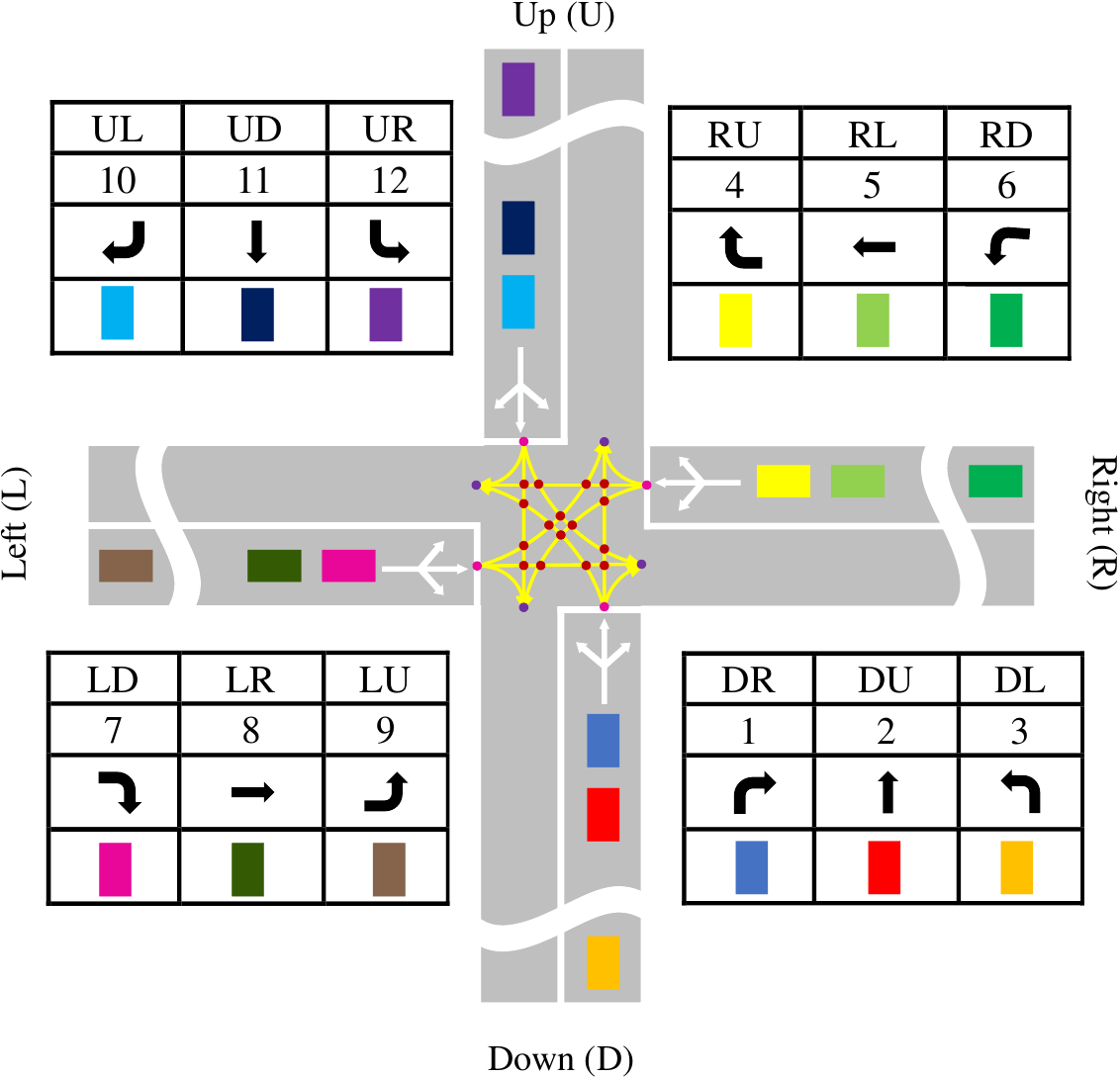}}
\caption{Intersection scenario settings}
\label{scenario}
\end{figure}

\begin{figure}[htbp]
\centerline{\includegraphics[width=0.8\linewidth]{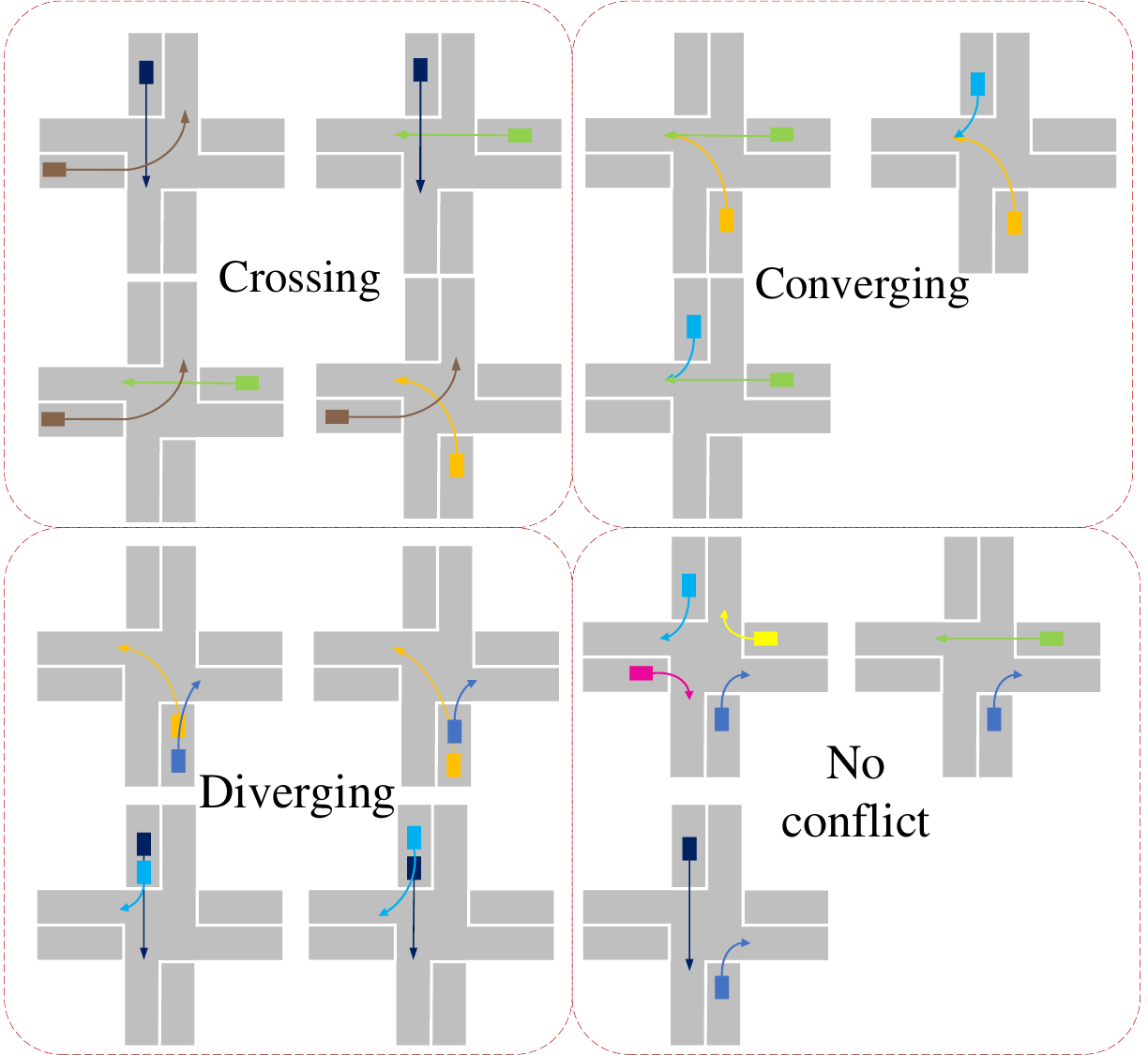}}
\caption{Typical modes of conflicts in the experiments}
\label{conflict modes}
\end{figure}

The scenario is a typical four-way single-lane intersection without traffic signals, as shown in Fig. \ref{scenario}. Each direction is denoted by its location in the figure, i.e. up (U), down (D), left (L) and right (R) respectively. We only focus on vehicles within a certain distance from the intersection center. Vehicles in each entrance are allowed to turn right, go straight or turn left. There are 12 types of vehicles, denoted by their entrances and exits, i.e. DR, DU, DL, RU, RL, RD, LD, LR, LU, UL, UD, and UR. Their number and meanings are listed in Table \ref{types of vehicle}. All their possible conflict relations are also illustrated in Fig.\ref{scenario}, which can be categorized into three classes, including crossing conflicts (denoted by red dots), converging conflicts (denoted by purple dots), and diverging conflicts (denoted by pink dots). To simplify our problem, we choose eight modes out of all the twelve modes to cover main conflicts. The rules are set as follows. First, each entrance includes two vehicles, in which one of them is a right-turning vehicle and the other is a vehicle turning left or going straight. Right-turning vehicles will only lead to converging and diverting conflicts, while vehicles going straight or turning left will introduce crossing conflicts. In the four entrances, two of them (entrances D and L) include a left-turning vehicle, and the other two (entrances U and R) include a vehicle going straight, which leads to a large number of crossing conflicts. In addition, the location and order of different vehicles in each entrance are random, which makes the problem still have enough complexity.
The eight modes include DR, DL, RU, RL, LD, LU, UL, UD, as shown in Fig. \ref{vehicles_in_experiment1}. From the figure, we can summary all types of conflicts it contains, which is shown in Fig. \ref{conflict modes}.
\begin{table}[b]
\captionsetup{justification=centering,labelsep=newline}
\caption{Different types of vehicles}
\small
\label{types of vehicle}
\vskip 0.15in
\begin{center}
\setlength{\tabcolsep}{4.5mm}{
\begin{tabular}{ccl}
\toprule
\textbf{Type} &\textbf{Number}  &\textbf{Meaning} \\
\midrule
DR &1 & From ‘Down’ turn right to ‘Right’ \\
DU &2 & From ‘Down’ go straight to ‘Up’\\
DL &3 & From ‘Down’ turn left to ‘Left’ \\
RU &4 & From ‘Right’ turn right to ‘Up’ \\
RL &5 & From ‘Right’ go straight to ‘Left’ \\
RD &6 & From ‘Right’ turn left to ‘Down’ \\
LD &7 & From ‘Left’ turn right to ‘Down’ \\
LR &8 & From ‘Left’ go straight to ‘Right’ \\
LU &9 & From ‘Left’ go straight to ‘Up’ \\
UL &10 & From ‘Up’ turn right to ‘Left’ \\
UD &11 & From ‘Up’ go straight to ‘Down’ \\
UR &12 & From ‘Up’ turn left to ‘Right’\\
\bottomrule
\end{tabular}}
\end{center}
\vskip -0.1in
\end{table}

We adopt the following assumptions. First, all vehicles are equipped with positioning and velocity devices so that we can gather location and movement information when they enter the interesting zone of the intersection. Second, only the longitudinal motion of the vehicles is controlled, assuming the vehicles follow the lateral trajectories. The environment model used in our MPC baseline and MA-PPO describing the longitudinal motion of vehicle $i$ is given by
\begin{equation}\label{eq.environment_model}
\begin{aligned}
    x^{\text{long}}_i(t+1) &= x^{\text{long}}_i(t) - v_i(t)\tau - \frac{1}{2}a_i(t)\tau^2\\
    v_i(t+1) &= v_i(t) + a_i(t)\tau,
\end{aligned}
\end{equation}
where $x^{\text{long}}_i$ is the longitudinal distance from the intersection, $v_i$ and $a_i$ are the velocity and acceleration, $\tau$ is the discrete time step. For the true dynamics of the simulation, the longitudinal motion $x^{\text{long}}_i$ is added with a Gaussian noise $\psi_i\sim\mathcal{N}(0, \sigma_i^2)$. The standard deviation $\sigma_i$ is related to the velocity, i.e., $\sigma_i = c_1 v_i \tau + c_2$, where $c_1$ controls the magnitude of the noise and $c_2$ is a small number to prevent numerical instability.
The problem is stated as follows: In each step, given positions and velocities of all the vehicles from the simulation, the acceleration of each vehicle needs to be decided in real time to let all vehicles pass the intersection safely and quickly.

\subsection{MPC and RL}
In this paper, we use VICS as a baseline for our comparison, which is a coordination scheme in the MPC framework. The scheme efficiently utilizes the intersection area by preventing each pair of conflicting vehicles from approaching their cross-collision point (CCP) at the same time, where the CCP is the intersection of their trajectories. A risk function is proposed to quantifying the risk of a collision of a pair of vehicles around their CCP, which is given by
\begin{equation}\label{eq.ccp}
    R_{i,j}(t)=H\delta_{i,j}e^{-(\alpha_i d_i^2+\alpha_j d_j^2)}.
\end{equation}
In this function, $H$ is a positive constant indicating the highest possible risk of collision, and $\delta_{i,j}$ is a binary variable to state whether the vehicles $i$ and $j$ have a CCP. Besides, $d_i$ and $d_j$ are the distances of the CCP from current position of vehicles $i$ and $j$ along their trajectories, and $\alpha_i$ and $\alpha_j$ are positive constants. At any time, if two conflicting vehicles are very close to their CCP, the risk function returns a high value, and if at least one vehicle is far from the CCP, it returns a low value. Based on that, a constrained nonlinear optimization problem is constructed as follows,
\begin{equation}\label{vics_formulation}
\begin{aligned}
J = &\sum_{t=0}^{T-1}\sum_{i=1}^{N}w_v(v_i(t+1)-v_d)^2+\\
&\quad\quad\quad\quad\sum_{t=0}^{T-1}\sum_{i=1}^{N}w_a(a_i(t))^2+\sum_{t=0}^{T-1}\sum_{i=1}^{N-1}\sum_{j=i+1}^{N}\mathcal{R}_{i,j}(t)\\
& \ \ \ \ \ \ \ \ \ \ \ \ s.t. \ \ v_{\text{min}} \le v_i \le v_{\text{max}},\\
&\ \ \ \ \ \ \ \ \ \ \ \ \ \ \ \ \ \  a_{\text{min}} \le a_i \le a_{\text{max}}
\end{aligned}
\end{equation}
where $T$ is the length of the prediction horizon, $v_d$ is the desired velocity, and $w_v$ and $w_a$ are weight coefficients. There are three cost terms in total. The first term denotes the cost related to velocity deviation from the desired value $v_d$. The second term denotes the cost of acceleration. Minimizing these two terms means comfortable and smooth flow of vehicles. The third term denotes the cost related to the risk of collisions as defined in the risk function \eqref{eq.ccp}, which sums up quantified risks at all CCPs for all possible pairs of vehicles considering their predicted trajectories in the horizon. Besides, constraints that are related to the velocity and acceleration limits are defined.

The constrained nonlinear optimization problem in the proposed MPC framework is solved in the SciPy package in python by the optimization toolbox \textit{minimize} using the Sequential Least Squares Programming (SLSQP). The prediction horizon is chosen as 20 steps. Detailed parameter settings are listed in Table \ref{Hyperparameters}.

Reinforcement learning comes from dynamic programming methods of optimal control, which is used to solve the optimal sequential decisions. Similar to MPC, which uses the model to minimize the sum of finite time-domain losses to optimize future sequential controls, reinforcement learning optimizes a policy to maximize the sum of future rewards. The difference is that MPC methods need to solve optimization problems online, while reinforcement learning can solve a policy offline and apply it online.

\subsection{RL formulation}
We are ready to transform our problem to an RL problem by defining state space, action space and reward function, which are basic elements in RL.
\subsubsection{State and action space}
By our assumption, we need to control at most eight vehicles at a time, i.e. two different types of vehicles at each entrance. Both the state and the action are defined as the concatenation of the state and action of each vehicle, i.e.,
\begin{equation}
\nonumber
\begin{aligned}
    s &= \left[s_{\text{DR}}, s_{\text{DL}}, s_{\text{RU}}, s_{\text{RL}}, s_{\text{LD}}, s_{\text{LU}}, s_{\text{UL}}, s_{\text{UD}}\right]\\
    a &= \left[a_{\text{DR}}, a_{\text{DL}}, a_{\text{RU}}, a_{\text{RL}}, a_{\text{LD}}, a_{\text{LU}}, a_{\text{UL}}, a_{\text{UD}} \right],
\end{aligned}
\end{equation}
where $s_{\text{*}}$ and $a_{\text{*}}$ denote the state and action of the vehicle type $\text{*}$.

\begin{figure}[htbp]
\centerline{\includegraphics[width=0.6\linewidth]{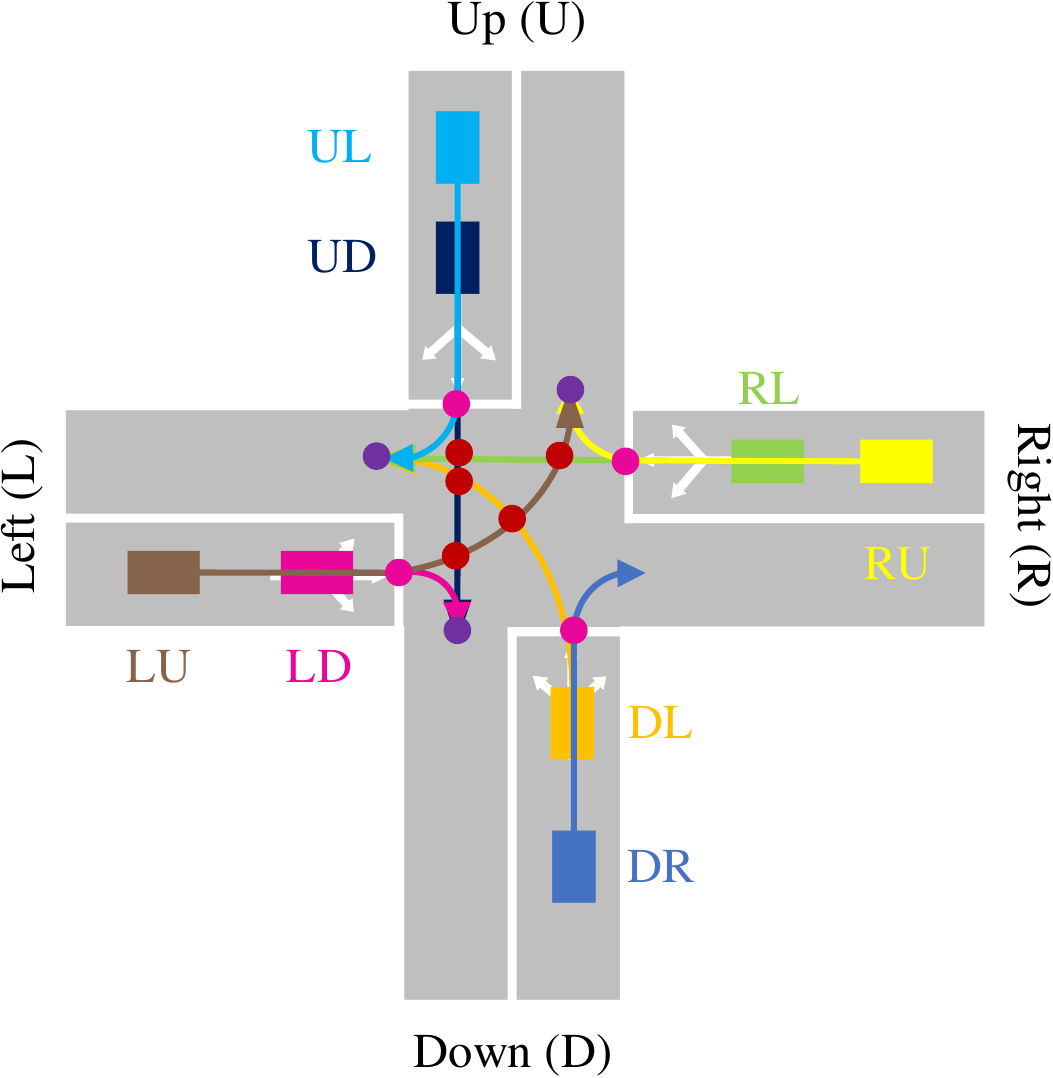}}
\caption{Vehicle modes chosen for the experiments}
\label{vehicles_in_experiment1}
\end{figure}

\begin{figure*}[tb]
\centerline{\includegraphics[width=0.8\linewidth]{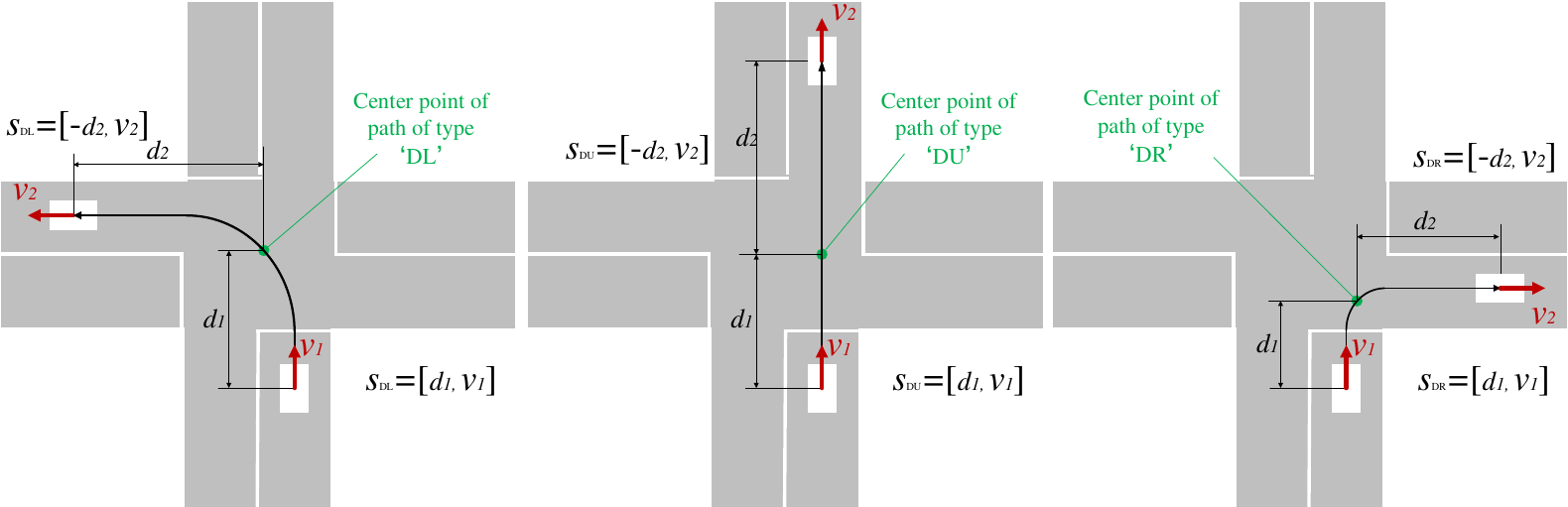}}
\caption{State formulation}
\label{state_formulation}
\end{figure*}

The state of each vehicle should contain position and velocity information. Intuitively, we can form the state by a tuple of coordinate and velocity, i.e. $(x, y, v)$, where $(x, y)$ is the coordinate of its position and $v$ is the velocity. However, by our task formulation, each vehicle has a fixed lateral trajectory corresponding to its type. There will be redundant information if we use this state formula. Besides, for continuous states, it is necessary to decrease the state space dimensionality to speed up learning and enhance stability. Observing all the trajectories are cross the intersection, we further compress the state of each vehicle by $(d, v)$, where $d$ is the distance between a vehicle and the center of its trajectory. Note that $d$ is positive when the vehicle is heading for the center and negative when it is leaving. The state formulation is shown as Fig. \ref{state_formulation}.

For the action, we use the acceleration of each vehicle. Finally, a 16-dimensional state space and an 8-dimensional action space are constructed.

\subsubsection{Reward settings}
Reward functions design concerns with convergence speed and asymptotic performance of RL algorithms. There are two points to note when defining reward functions. First, the complexity of the reward function and the convergence speed should be balanced. The more detailed the definition of the reward function, the faster the algorithm will learn, but defining such a reward function will require a lot of human design, especially in high-dimensional state space and action space. Second, positive and negative rewards have different functionality. Positive rewards drive the system to avoid terminals to maximize accumulating rewards unless the terminal yields a large positive reward, while negative rewards incentive the agent to reach a terminal state as quickly as possible to avoid accumulating penalties.

For our high-dimensional RL problem, a look-up table is adopted as the reward function, which is much easier to design compared with a continuous function.
The reward function is designed under consideration of safety, efficiency and task completion. First of all, the task is designed in an episodic manner, in which a bad terminal state and a good terminal state are given, i.e., collision and all vehicles
passing the intersection. In order to learn a safe policy, when the system reaches the bad terminal state (collision), we set up a large negative reward to discourage its presence. In order to improve traffic efficiency, we set a small negative reward at each step to drive the system to the good terminal state quickly. To encourage task completion, we set a positive reward as long as some vehicle passes the intersection and a large positive reward when the good terminal state is reached (all vehicles passing the intersection). All reward settings are listed in Table \ref{reward}.

\begin{table}[t]
\captionsetup{justification=centering,labelsep=newline,font=small}
\caption{Reward settings}
\small
\label{reward}
\vskip 0.15in
\begin{center}
\setlength{\tabcolsep}{10mm}{
\begin{tabular}{lc}
\toprule
\textbf{Reward items}  &\textbf{Reward}  \\
\midrule
Collision  &-50 \\
Step reward  &-1 \\
Some vehicle passes  &10 \\
All vehicles pass  &50 \\
\bottomrule
\end{tabular}}
\end{center}
\vskip -0.1in
\end{table}

\subsection{Algorithm architecture}
\begin{figure}[htbp]
\centerline{\includegraphics[width=0.87\linewidth]{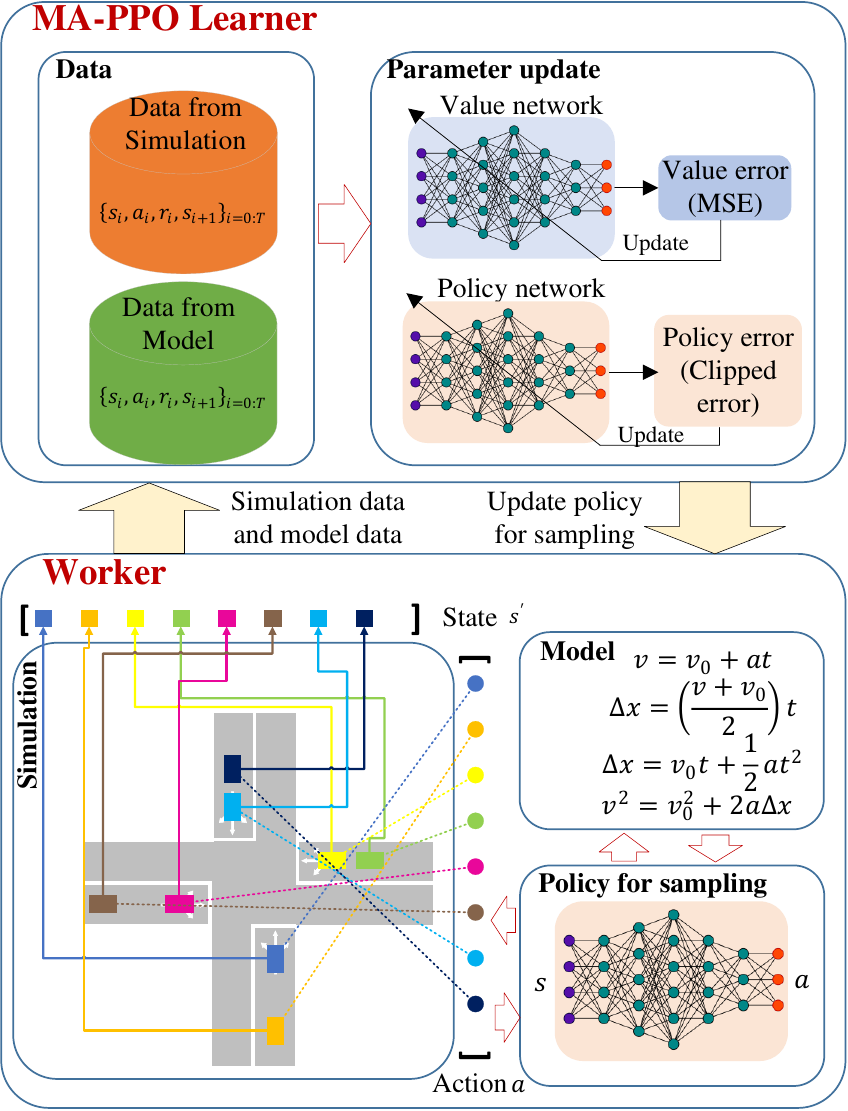}}
\caption{Overall architecture of MA-PPO}
\label{overall_architecture}
\end{figure}

In this section, we illustrate how to apply MA-PPO algorithm to this centralized coordination problem.
The MA-PPO architecture consists of two main parts, including MA-PPO learner and worker. The worker is in charge of getting the updated policy from the learner and using it to collect experience data from simulation and the environment model. MA-PPO learner then uses the data collected from the worker to update the value and policy networks by gradient descent methods, and finally syncs the updated parameters to the worker for the next iteration. The overall architecture is shown in Fig. \ref{overall_architecture}.

\section{Experiments}\label{Experiment settings}
\subsection{Experimental settings}
Four sets of experiments are conducted in this section. In the first experiment, we train a policy by MA-PPO and compare it with a policy from the early training stage to show improvements over the training process. The second experiment compares computing time and traffic efficiency of the policy trained by MA-PPO with that of VICS. The third experiment illustrates the training process of MA-PPO and PPO, which shows our algorithm has better sample efficiency. The fourth experiment shows the performance comparison between MA-PPO and PPO under different environment noise, and shows that we can control the performance loss caused by the model inaccuracy through adjusting the depth of model rollouts in MA-PPO. All the experiments are carried out in a simulation environment, in which the scenario is shown in Fig. \ref{vehicles_in_experiment1}. The initial positions of all vehicles are random. Considering it is a small intersection in our experiments, we only verify our algorithm in a low-speed scene. The maximum speed of all vehicles is limited to 10 m/s.

Detailed parameter settings of all these experiments are listed in Table \ref{Hyperparameters}. The time step of all simulation experiments is set to be 0.1s, and the number of vehicles is set to be 8. Except for the fourth experiment, the environment noise related parameters $c_1$ and $c_2$ are set to be 1/30 and 2e-7 respectively, which are defined in the Section \ref{problem statement}. We employ multiple layers percepton with two hidden layers as the approximate functions of the policy and the value function. Both of them have 128 units in each hidden layer, And the policy network has 16 output units to parameterize the Gaussian distribution of the accelerations of all vehicles, while the value function has only one output unit for the state value. In each iteration, each worker collects $B$=2048 transitions and uses minibatch size $MB$=64 and epoch $U$=10 for updating. To stabilize the training process, the learning rate $LR$ is set to linearly decrease from 0.0003 to 0, and Adam is selected as the optimizer to stabilize the gradient direction. The training process is not terminated until the number of collected real samples reaches 5e7. Sixteen parallel workers are used to improve exploration and stabilize the learning process. In each iteration, each worker collects a batch of data, then takes the first minibatch to calculate the gradient, then the global gradient is obtained by averaging all local gradients of workers. Each worker updates its parameters using the global gradient and goes on like this. Rollout depth $D$ is only a parameter of MA-PPO, which is used to control the depth of model rollouts. Except for the fourth experiment, we set it to be 50. The second experiment involves the comparison of MA-PPO and VICS. The parameters of VICS refer to the settings in its original paper. The prediction horizon is set to be 20 steps, and the vehicle speed is limited to the interval [0, 10m/s], and the acceleration is limited to the interval [-5$\rm m/s^2$, 5$\rm m/s^2$]. The desired velocity $v_d$ is set to 8m/s. Besides, the weights of the velocity term and the acceleration term in the objective function is $w_v$=1 and $w_a$=5 respectively, and the parameters of the risk term are $H$=1000, $\alpha$=0.005.

\subsection{Comparison of results in different training stages}
In this experiment, we train a policy by MA-PPO and then compare it with a policy from the early training stage.

\begin{table}[t]
\captionsetup{justification=centering,labelsep=newline,font=small}
\caption{Hyperparameters of the experiments}
\small
\centering
\label{Hyperparameters}
\vskip 0.15in
\begin{threeparttable}
\begin{tabular}{lc}
\toprule
\textbf{Parameters}&\textbf{Value} \\
\midrule
\textit{Simulator}&\\
\quad Discrete time step $\tau$ & 0.1s\\
\quad Vehicle numbers $N$ & 8\\
\quad Simulation noise  & $c_1$=1/30, $c_2$=2e-7\\

\midrule
\textit{MA-PPO \& PPO}&\\
\quad Discount factor $\gamma$ & 0.99 \\
\quad $\lambda$ & 0.95 \\
\quad Clip range $\epsilon$ & 0.2 \\
\quad Total timesteps $T_{total}$ & 5e7  \\
\quad Seed number & 5 \\
\quad Rollout depth $D$ \tnote{*} & 50\\
\quad Batch size $B$ & 2048 \\
\quad Minibatch size $MB$ & 64 \\
\quad Epoch $U$ & 10  \\
\quad Learning rate $LR$ & 0.0003 $\rightarrow$ 0 \\
\quad Hidden layer number & 2 \\
\quad Hidden units number & 128 \\
\quad Optimizer & Adam  \\
\quad Number of workers& 16\\
\midrule
\textit{VICS}&\\
\quad Predictive horizon $T$ & 20\\
\quad Maximum velocity $v_{\text{max}}$ & 10${\rm m/s}$\\
\quad Minimum velocity $v_{\text{min}}$ & 0${\rm m/s}$ \\
\quad Maximum acceleration $a_{\text{max}}$ & 5${\rm m/s^2}$\\
\quad  Minimum acceleration $a_{\text{min}}$ & -5${\rm m/s^2}$\\
\quad Desired velocity $v_d$ & 8${\rm m/s}$\\
\quad $w_v$ & 1\\
\quad $w_a$ & 5\\
\quad $H$ & 1000\\
\quad $\alpha$ & 0.005\\
\bottomrule
\end{tabular}
\begin{tablenotes}
     \item[*] Parameter used only by MA-PPO
\end{tablenotes}
\end{threeparttable}
\vskip -0.1in
\end{table}

\begin{figure}[htbp]
\centering
\captionsetup[subfigure]{justification=centering}
\subfloat[An episode with collision]{\label{fail1}\includegraphics[width=0.23\textwidth]{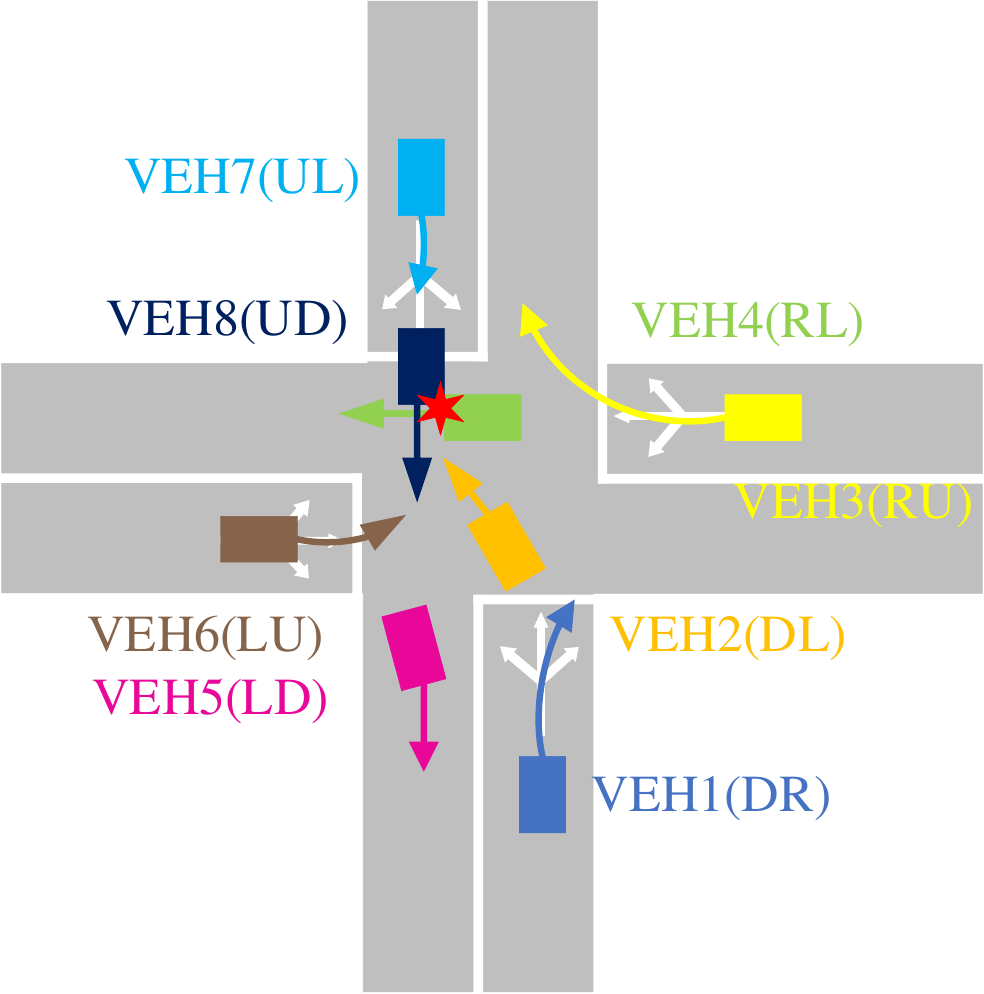}} 
    \subfloat[Distance vs Timestep]{\label{fail2}\includegraphics[width=0.23\textwidth]{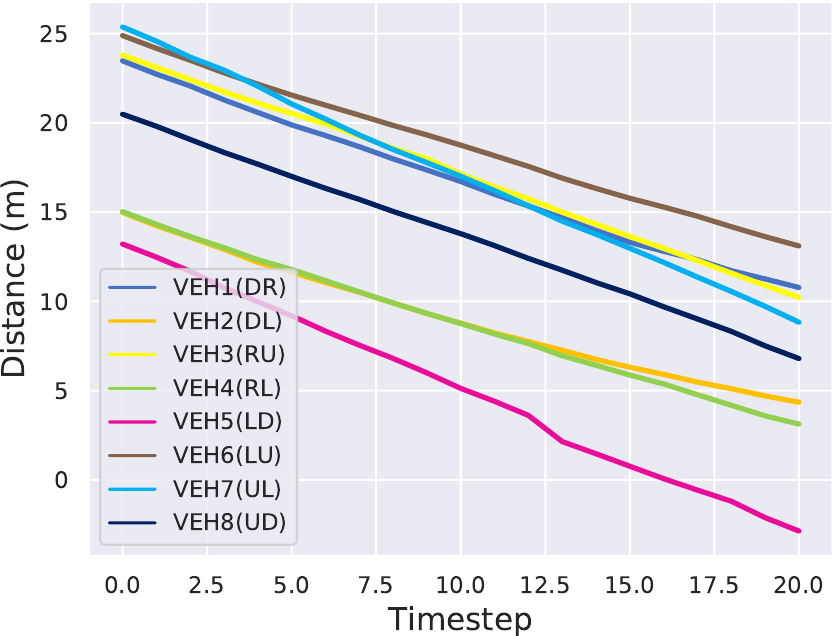}} \\
    \subfloat[Velocity vs Timestep]{\label{fail3}\includegraphics[width=0.23\textwidth]{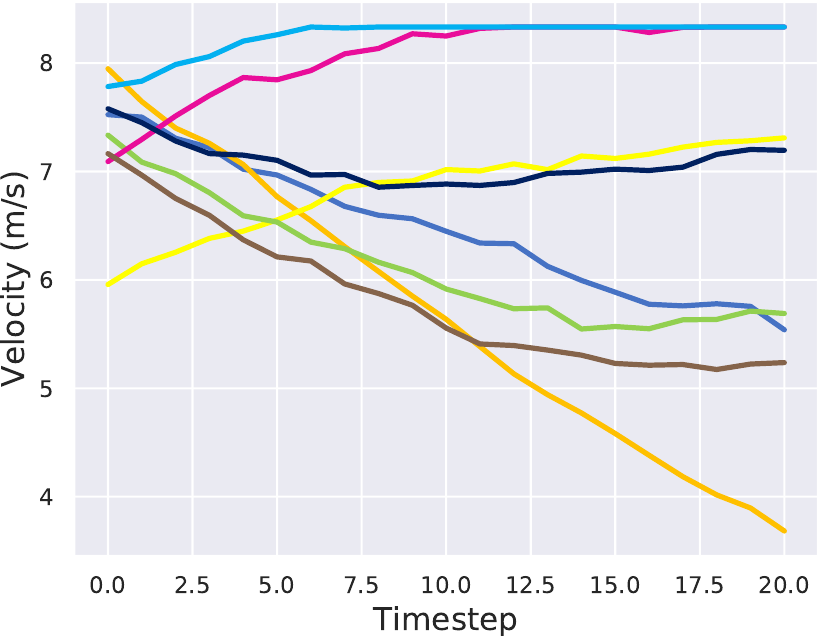}}
    \subfloat[Action vs Timestep]{\label{fail4}\includegraphics[width=0.23\textwidth]{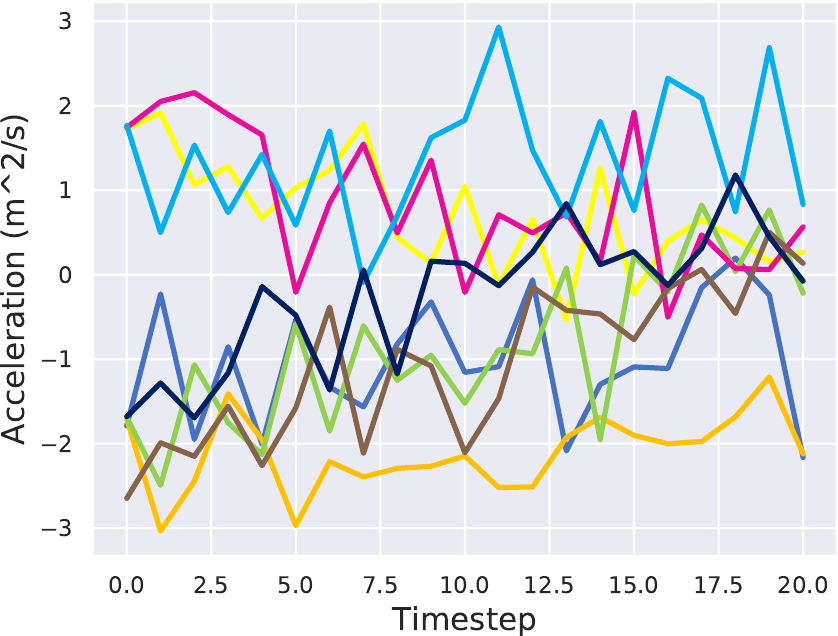}} \\ 
\caption{Results from the 20th iteration of the training process, including the distance to the intersection center, the speed information and the action they take during the episode}
\label{fig:fail}
\end{figure}

Fig. \ref{fig:fail} visualizes an episode in the 20th iteration of the training process. In this episode, VEH5 (mode: LD) passes the intersection successfully. However, VEH4 (mode: RL) and VEH8 (mode: UD) collide.  
From Fig. \ref{fail2} we can see all the eight vehicles are approaching the center of the intersection, but almost none of them realize to decelerate to avoid collision except for VEH2.
During the last few steps before the collision, the velocities of VEH4 and VEH8 still maintain their trend without significant changes. The random curves in the acceleration illustrated in Fig \ref{fail4} also show that the learned policy at this point fails to coordinate all the vehicles.

\begin{figure}[htbp]
\centering
\captionsetup[subfigure]{justification=centering}
\subfloat[An episode without collision]{\label{success1}\includegraphics[width=0.23\textwidth]{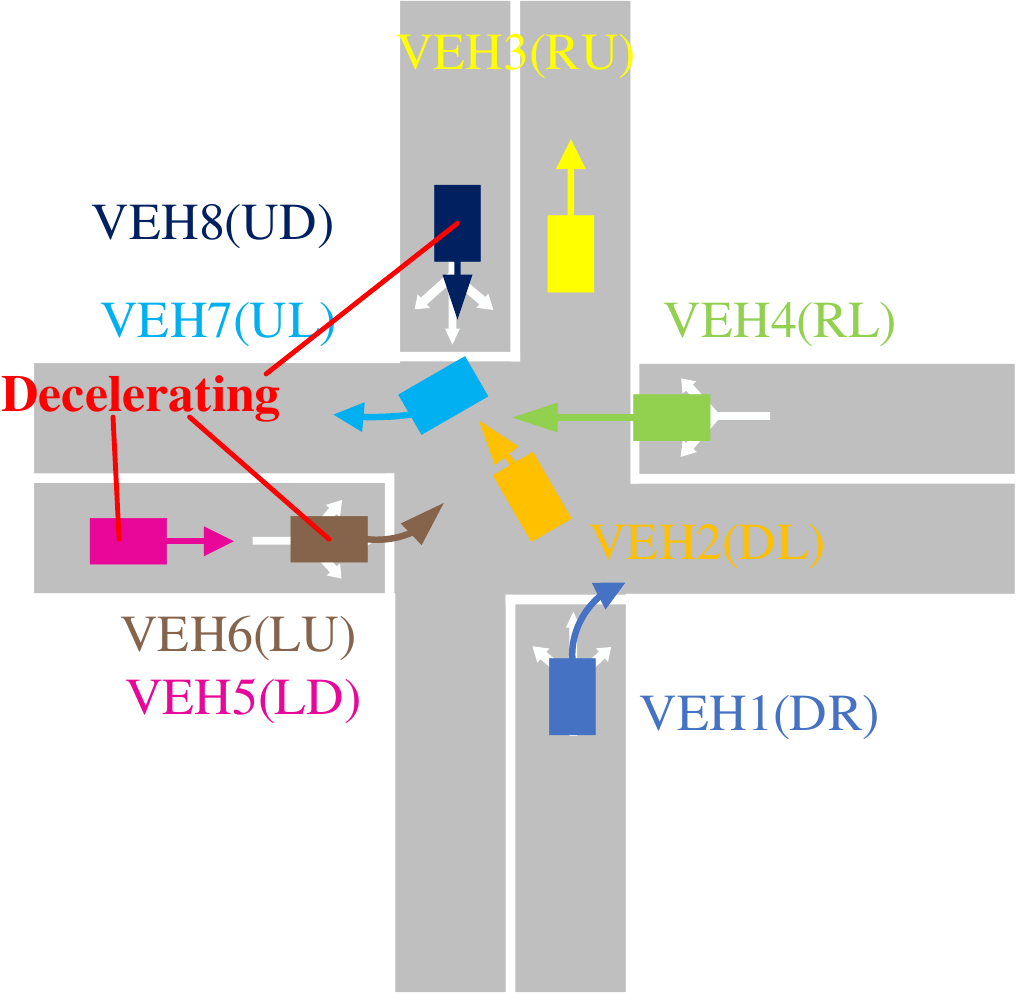}} 
    \subfloat[Distance vs Timestep]{\label{success2}\includegraphics[width=0.23\textwidth]{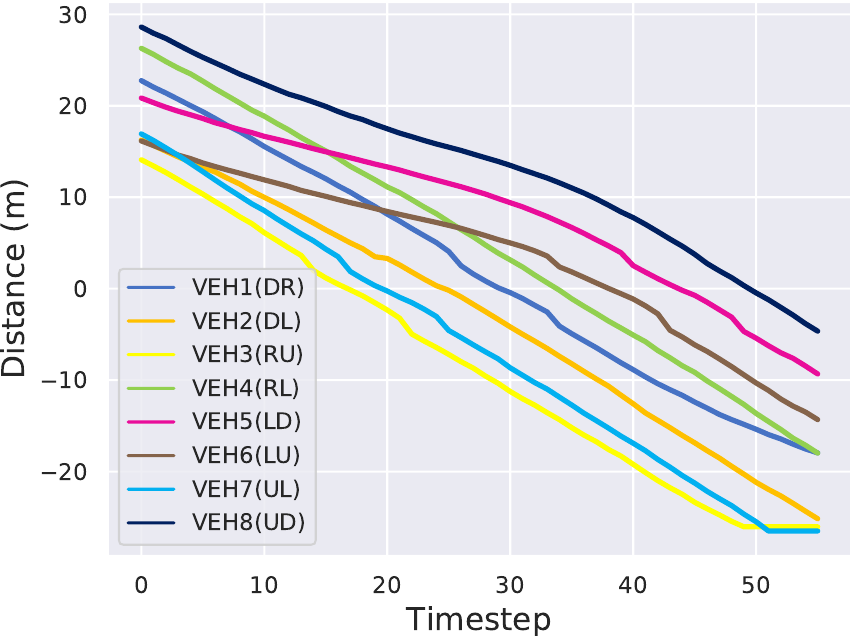}} \\
    \subfloat[Velocity vs Timestep]{\label{success3}\includegraphics[width=0.23\textwidth]{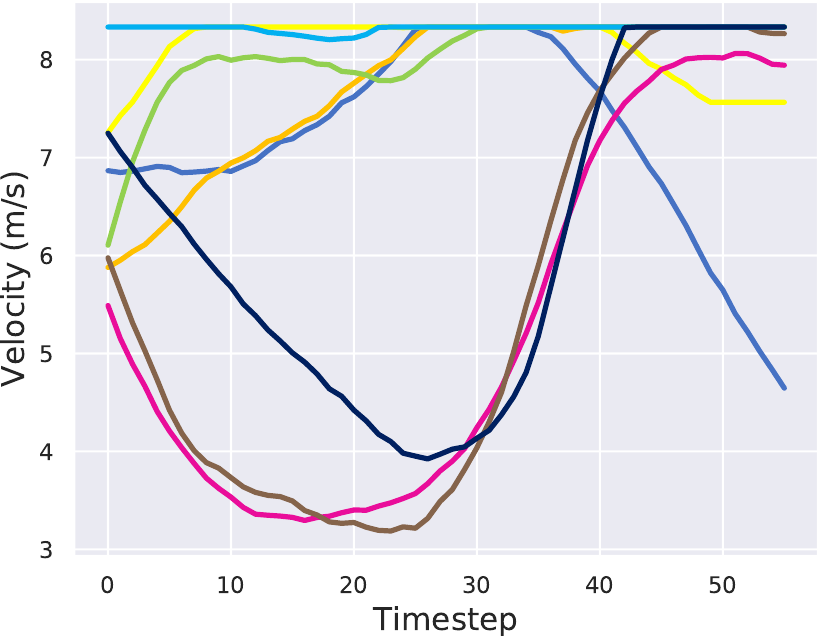}}
    \subfloat[Action vs Timestep]{\label{success4}\includegraphics[width=0.23\textwidth]{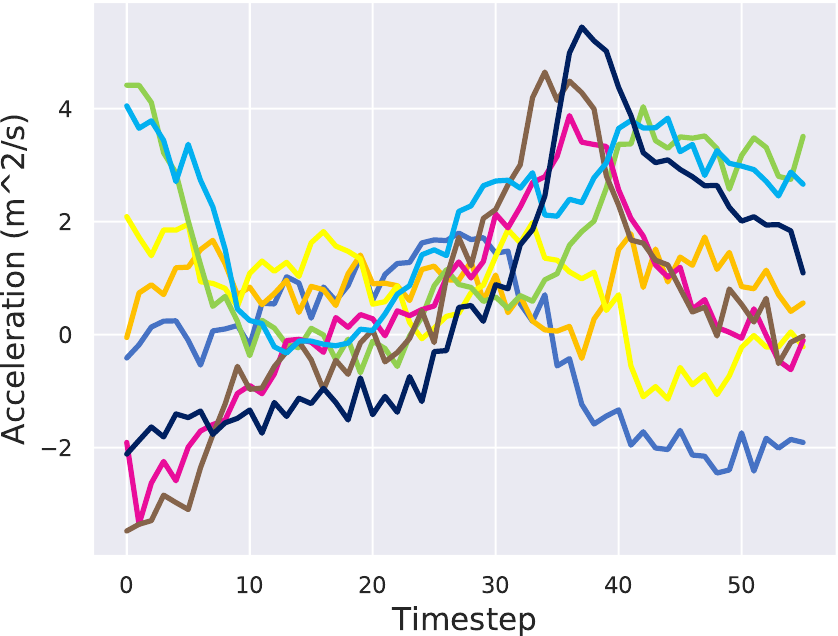}} \\ 
\caption{Results from the 500th iteration of the training process, including the distance to the intersection center, the speed information and the action they take during the episode}
\label{fig:success}
\end{figure}

Fig. \ref{fig:success} illustrates the result after 500 iterations of training, which shows a good coordination scheme has been learned. In this episode, VEH3 (mode: RU) first passes the intersection. VEH8 slows down before the timestep 26 to wait for VEH7 to turning right. Besides, VEH2 (mode: DL) keeps a low speed to wait for the pass of VEH7. Also, VEH4 (mode: RL) decelerates to wait for VEH2 turning left first. One reasonable explanation that VEH8 has to wait and pass lastly is that it has a longer distance to the center of the intersection than any other vehicles, as shown in Fig. \ref{success2}. It can be seen from Fig. \ref{success4} that, VEH8 remains stable deceleration between -2$\rm m/s^2$ and -1$\rm m/s^2$ until the timestep 26, when VEH7, VEH2 and VEH4 have passed the central area of the intersection.

VEH5 and VEH6 have similar velocity curves. In the beginning, they slow down and keep low velocity until VEH2, VEH3 and VEH4 pass the intersection. After the timestep 25, both of them begin to speed up and pass the intersection because there is no risk of collision around the center of the intersection. On the other hand, the velocity curves of VEH2, VEH3 and VEH4 demonstrate that they tend to keep a high speed so that they could pass the intersection quickly. To sum up, Fig. \ref{fig:success} shows that MA-PPO has learned a human-like policy in the 500th iteration, which can avoid the potential collision, and coordinate all the vehicles passing the intersection efficiently.

\subsection{Comparison of MA-PPO policy and VICS}
In this experiment, we implement VICS in our experiment and compare traffic efficiency and computing time with the policy trained by MA-PPO.

For the implementation of VICS, we follow the method in the original paper. First, we define the CCP for each pair of conflicting vehicles in our scenario. Then we formulate the nonlinear optimization problem \eqref{vics_formulation} with coefficients shown in Table \ref{Hyperparameters}. All evaluations were performed on a single computer with a 2.9GHz Intel(R) Core(TM) i9-8950HK CPU.

We use the average episode length and the average computing time as metrics of traffic and computing efficiency respectively. The average episode length is the average number of timesteps over 10 episodes, where an episode means the process from initialization to all vehicles passing the intersection. The average computing time is the average calculation time spent in all timesteps over 10 episodes. Besides, we also compute the variance of the computing time. It is obvious to see that the shorter the average episode length, the higher the traffic efficiency, and the less the average computing time, the lower the overhead. The results are shown in Table \ref{compare_with_mpc}. From the results, it can be seen that the traffic efficiency of our method is 4.5 times that of VICS and our method is around 400 times more efficient than VICS in terms of computing efficiency. Besides, the variance of computing time of VICS is much higher than that of our method, which means the computing time of VICS varies in different timesteps. Actually, the maximum value is up to 30s in our experiments, which makes it not practical to be applied in the real world. On the other hand, our method can meet the real time requirements while it has better performance.

\begin{table}[t]
\captionsetup{justification=centering,labelsep=newline,font=small}
\caption{Comparison of our method and VICS}
\small
\label{compare_with_mpc}
\vskip 0.15in
\begin{center}
\setlength{\tabcolsep}{4mm}{
\begin{tabular}{lcc}
\toprule
&\textbf{Ours}&\textbf{VICS} \\
\midrule
Average episode length & 53.44 &  241.0 \\
Average computing time &0.00417s & 1.64560s \\
Variance of computing time & 0.000247 &13.657 \\
\bottomrule
\end{tabular}}
\end{center}
\vskip -0.1in
\end{table}

\begin{figure}[htbp]
\centering
\captionsetup[subfigure]{justification=centering}
\subfloat[An episode using VICS]{\label{vics1}\includegraphics[width=0.23\textwidth]{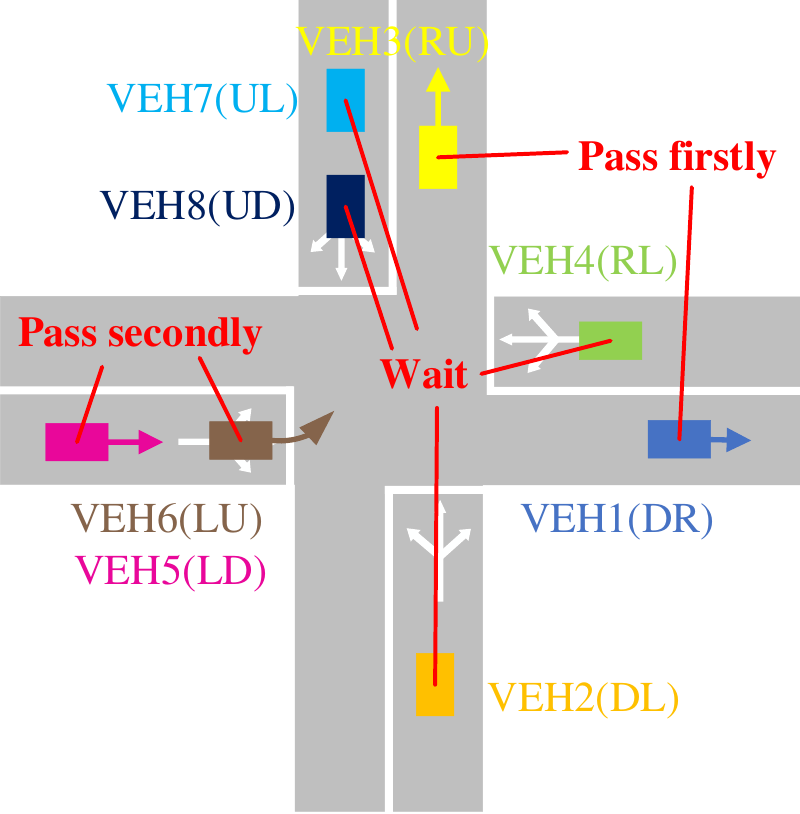}} 
    \subfloat[Distance vs Timestep]{\label{vics2}\includegraphics[width=0.23\textwidth]{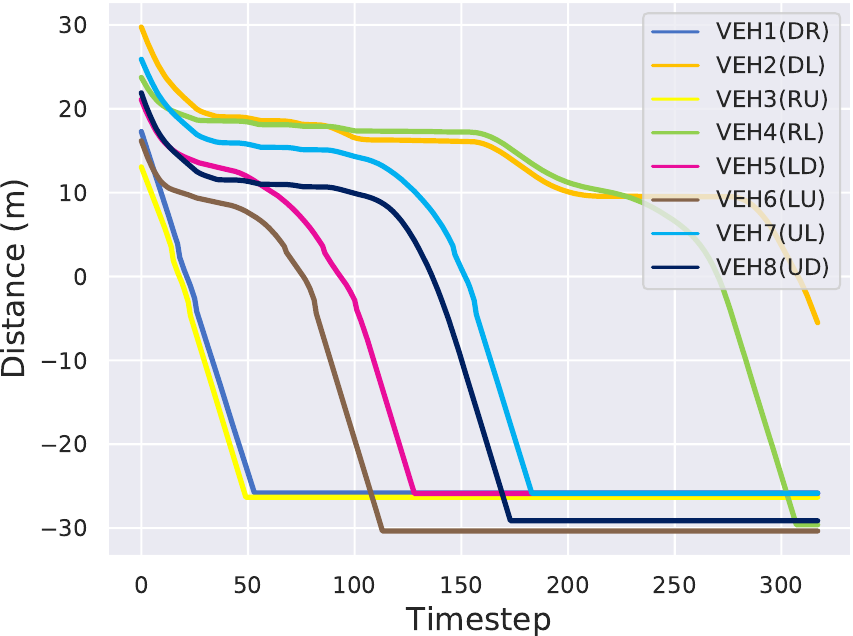}} \\
    \subfloat[Velocity vs Timestep]{\label{vics3}\includegraphics[width=0.23\textwidth]{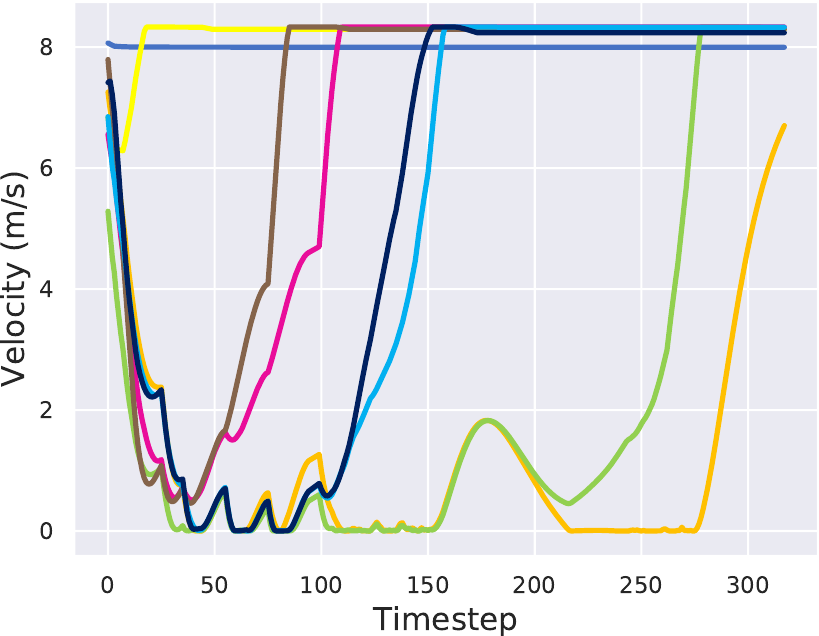}}
    \subfloat[Action vs Timestep]{\label{vics4}\includegraphics[width=0.23\textwidth]{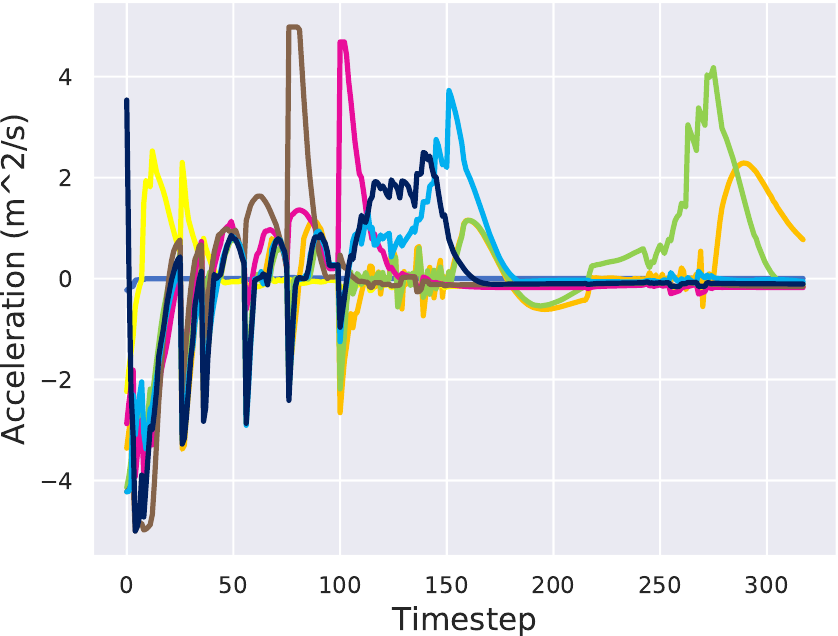}} \\ 
\caption{A typical episode of VICS, including the distance to the intersection center, the speed information and the action taken during the episode}
\label{fig:mpc}
\end{figure}

We also show a typical episode of VICS in Fig. \ref{fig:mpc}. Comparing it with Fig. \ref{fig:success}, we can see that VICS tends to stop all the vehicles which have conflict relationship with the vehicles in the intersection. It lets VEH1 (mode: DR) and VEH3 (mode: RU) cross the intersection first, then VEH5 (mode: LD) and VEH6 (mode: LU) accelerate to pass, followed by VEH7 (mode: UL) and VEH8 (mode: UD), and finally VEH2 (mode: DL) and VEH4 (mode: RL) pass the intersection. On the other hand, our method only slows down conflict vehicles to a lower speed rather than makes it pull up. As a result, VICS is less efficient than our approach.

\subsection{Sample efficiency of MA-PPO and PPO}
\begin{figure}[htbp]
\centering
\captionsetup[subfigure]{justification=centering}
\subfloat[Mean episode reward]{\label{epreward}\includegraphics[width=0.24\textwidth]{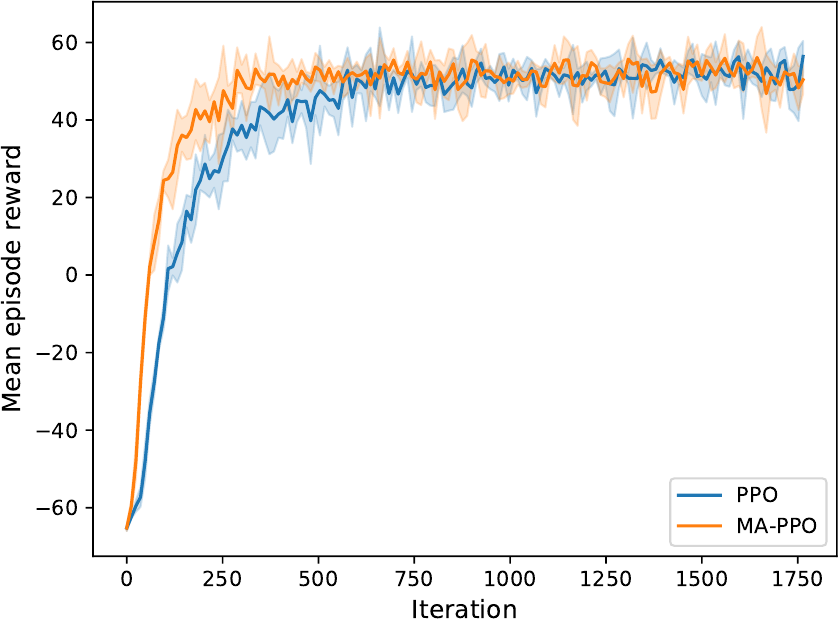}} 
\subfloat[Mean episode length]{\label{eplength}\includegraphics[width=0.23\textwidth]{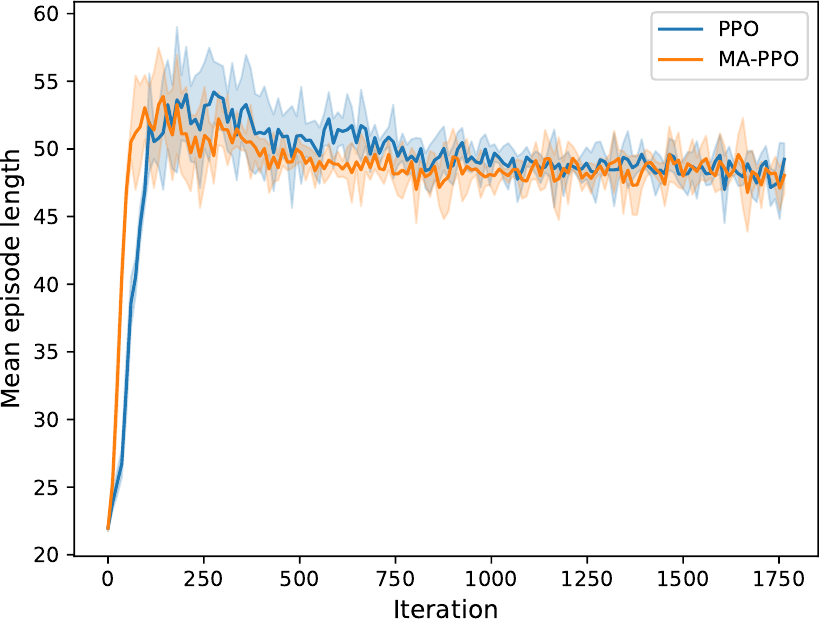}} 
\caption{Training process of MA-PPO and PPO}
\label{process}
\end{figure}

In the third experiment, we use MA-PPO and PPO to train a policy respectively and illustrate their training processes to show the superiority of our proposed algorithm. To eliminate the impact of randomness, we train both PPO and MA-PPO under five random seeds. The results are shown in Fig. \ref{process}, where the solid curve represents the mean value and the shaded region represents the variance over different random seeds.

Fig. \ref{epreward} shows the mean episode reward of MA-PPO and PPO during the training process. Both MA-PPO and PPO get the highest reward around 50, which means that all the eight vehicles pass the intersection successfully. Compared with PPO, MA-PPO converges at around 500 iterations, while PPO algorithm needs about 1000 iterations, which shows that MA-PPO converges almost twice as fast as PPO.  

Fig. \ref{eplength} shows the change of the mean episode length during the training process. The episode lengths of MA-PPO and PPO first increase rapidly and then reduce to an equal value. This can be explained that at the beginning, the learned policy mainly focuses on how to avoid colliding because this will yield a large negative reward. At that time, the vehicles tend to wait or drive in a very low speed until there is no risk of colliding, which leads to the long episode length. However, such a policy is too conservative and suffers low efficiency. Therefore, the following policy would optimize this process to avoid long waiting time, leading to the decrease of the mean episode length. Similar to Fig. \ref{epreward}, MA-PPO obtains faster convergence speed in term of the mean episode length compared with PPO.

\subsection{Asymptotic performance under different environment noise}
\begin{figure}[htbp]
\centerline{\includegraphics[width=0.7\linewidth]{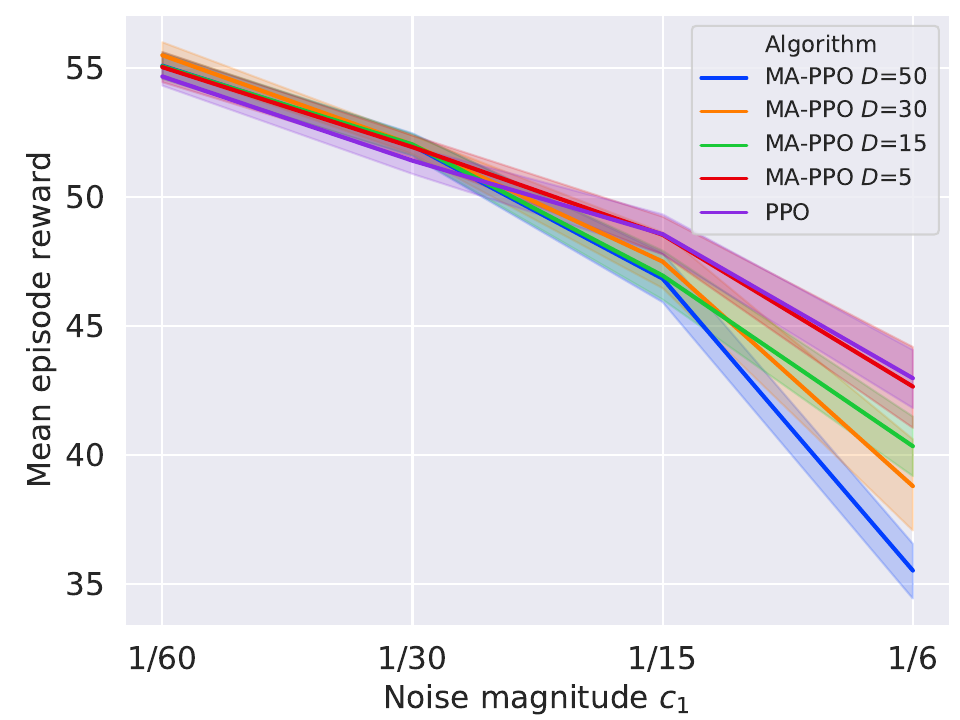}}
\caption{Asymptotic performance under different environment noise}
\label{fig.noise}
\end{figure}

In this experiment, we will explore the influence of different environment noise on the asymptotic performance of PPO and MA-PPO algorithm. As shown in Section \ref{problem statement}, the noise in the simulation environment is a Gaussian noise with zero mean value. We adjust the noise by controlling its standard deviation, specifically, $c_1$. For MA-PPO, the model in \eqref{eq.environment_model} is used. With the increase of noise in the environment, the inaccuracy of the model is also increasing. Compared with PPO algorithm, learning with virtual samples will inevitably lead to performance decay. We explore the influence of the hyperparameter $D$ in MA-PPO algorithm on the performance, which is the depth of model rollouts. In the experiment, we take $c_1$ as 1/60, 1/30, 1/15 and 1/6 to get different environment noise. In addition, the hyperparameter $D$ of MA-PPO algorithm is set to be 50, 30, 15 and 5 respectively. For each group of parameters, we train a policy and run it 15 episodes respectively to obtain the mean episode reward, which is used as the asymptotic performance. The results are shown in Fig. \ref{fig.noise}.

It can be seen from the figure that for both of PPO and MA-PPO, the asymptotic performance decreases with the increase of noise in the simulation environment, and its variance is also gradually increasing. This is because the uncertainty of the simulation environment increases, as a result, the learned policies tend to be conservative, and the total time of crossing the intersection becomes longer, so the sum of the rewards will also decrease accordingly. In addition, MA-PPO is more robust to small Gaussian errors of the model, for example, when $c_1$ equals 1/60 or 1/30, no matter which value of $D$ is taken, the performance of MA-PPO is not significantly reduced compared with PPO algorithm. However, when the model error is large ($c_1$=1/15 or 1/6), the larger the $D$, the worse the asymptotic performance of MA-PPO. This is because when the model error is large, the uncertainty and inaccuracy of the virtual samples will grow with the increase of rollout depth, which will lead to the rise in the error of the value function and the policy gradient, and finally influence the final performance. Therefore, the hyperparameter $D$ can be adjusted according to the model error in practice to acquire good results on both the sample efficiency and the asymptotic performance.

\section{Conclusion}\label{Conclusion}
In this paper, we propose a centralized coordination scheme of automated vehicles at an intersection without traffic signals using reinforcement learning (RL) to address low computation efficiency suffered by current centralized coordination methods, which have been long regarded as a challenging problem due to the real time requirement. We first propose an RL training algorithm, model accelerated proximal policy optimization (MA-PPO), which incorporates a prior model into proximal policy optimization (PPO) algorithm to enhance sample efficiency. Then we present the design of state, action and reward in a typical four-way single-lane intersection which contains eight different modes of vehicles to formulate centralized coordination as an RL problem. Finally, a coordination policy is trained and evaluated through numerical simulation. It is observed that a human-like policy with high traffic efficiency is got at the end of the training process. We compare its performance with VICS, a coordination scheme based on model predictive control method. Results show that our method spends only 1/400 of the computing time of VICS and increases the efficiency of the intersection by 4.5 times. Besides, we also show that by using a prior model, MA-PPO speeds up the learning process by two times, which shows the superiority of the proposed algorithm.

But our work still needs to be improved. First of all, the trained policy cannot guarantee the safety in real-testing, although there is a strict collision penalty in the reward design and the policy does work well in our experiments. We cannot prove theoretically that our trained policy is collision-free. 
Second, we only consider the longitudinal dynamics of vehicles while the lateral and longitudinal dynamics are coupled in the real world, which makes the stability of vehicles unable to be guaranteed. In order to solve these problems, we will introduce safety constraints and stability constraints into the original problems in the future work. Since the carrier of our policy is a neural network with a large number of parameters, we will focus on how to solve such a large-scale constrained reinforcement learning problem, such as using penalty function methods or gradient projection methods.

\section{Acknowledgments}\label{sec11}
This work is partially supported by International Science $\&$ Technology Cooperation Program of China under 2016YFE0102200 and Tsinghua University-Toyota Joint Research Center for AI Technology of Automated Vehicle. We would like to acknowledge Mr. Jingliang Duan, Mr. Zhengyu Liu, for their valuable suggestions throughout this research.

\ifCLASSOPTIONcaptionsoff
  \newpage
\fi

\bibliographystyle{ieeetr}
\bibliography{bare_jrnl.bbl}

\begin{IEEEbiography}[{\includegraphics[width=1in,height=1.25in,clip,keepaspectratio]{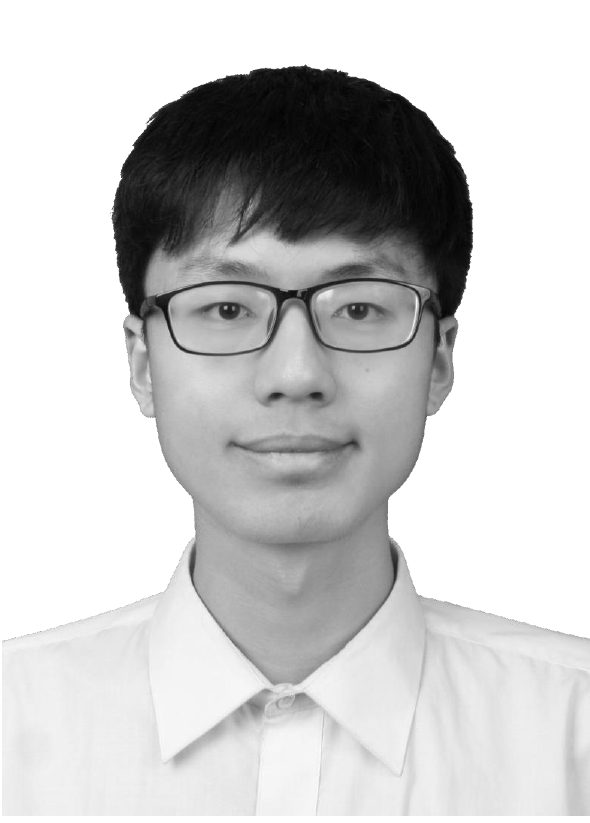}}]{Yang Guan}
received the B.S. degree from school of mechanical engineering, Beijing institute of technology, Beijing, China, in 2017. He is pursuing his Ph.D. degree in the School  of  Vehicle  and  Mobility, Tsinghua University, Beijing, China.  His research interests include decision-making of autonomous vehicle, and reinforcement learning.
\end{IEEEbiography}
\vskip -2\baselineskip plus -1fil

\begin{IEEEbiography}[{\includegraphics[width=1in,height=1.25in,clip,keepaspectratio]{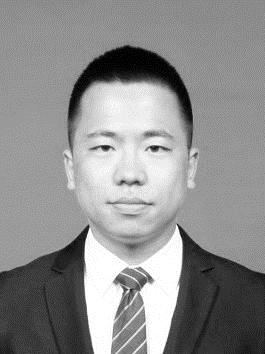}}]{Yangang Ren}
received the B.S. degree from the Department of Automotive Engineering, Tsinghua University, Beijing, China, in 2018. He is pursuing his Ph.D. degree in the School of Vehicle and Mobility, Tsinghua University, Beijing, China. His research interests include decision and control of autonomous driving, reinforcement learning and adversarial learning.
\end{IEEEbiography}

\vskip -2\baselineskip plus -1fil

\begin{IEEEbiography}[{\includegraphics[width=1in,height=1.25in,clip,keepaspectratio]{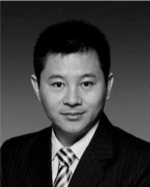}}]{Shengbo Eben Li}
(SM'16) received the M.S. and Ph.D. degrees from Tsinghua University in 2006 and 2009. He worked at Stanford University, University of Michigan, and University of California, Berkeley. He is currently a tenured associate professor at Tsinghua University. His active research interests include intelligent vehicles and driver assistance, reinforcement learning and distributed control, optimal control and estimation, etc.

He is the author of over 100 journal/conference papers, and the co-inventor of over 20 Chinese patents. He was the recipient of Best Paper Award in 2014 IEEE ITS Symposium, Best Paper Award in 14th ITS Asia Pacific Forum, National Award for Technological Invention in China (2013), Excellent Young Scholar of NSF China (2016), Young Professorship of Changjiang Scholar Program (2016). He is now the IEEE senior member and serves as associated editor of IEEE ITSM and IEEE Trans. ITS, etc.
\end{IEEEbiography}

\vskip -2\baselineskip plus -1fil

\begin{IEEEbiography}[{\includegraphics[width=1in,height=1.25in,clip,keepaspectratio]{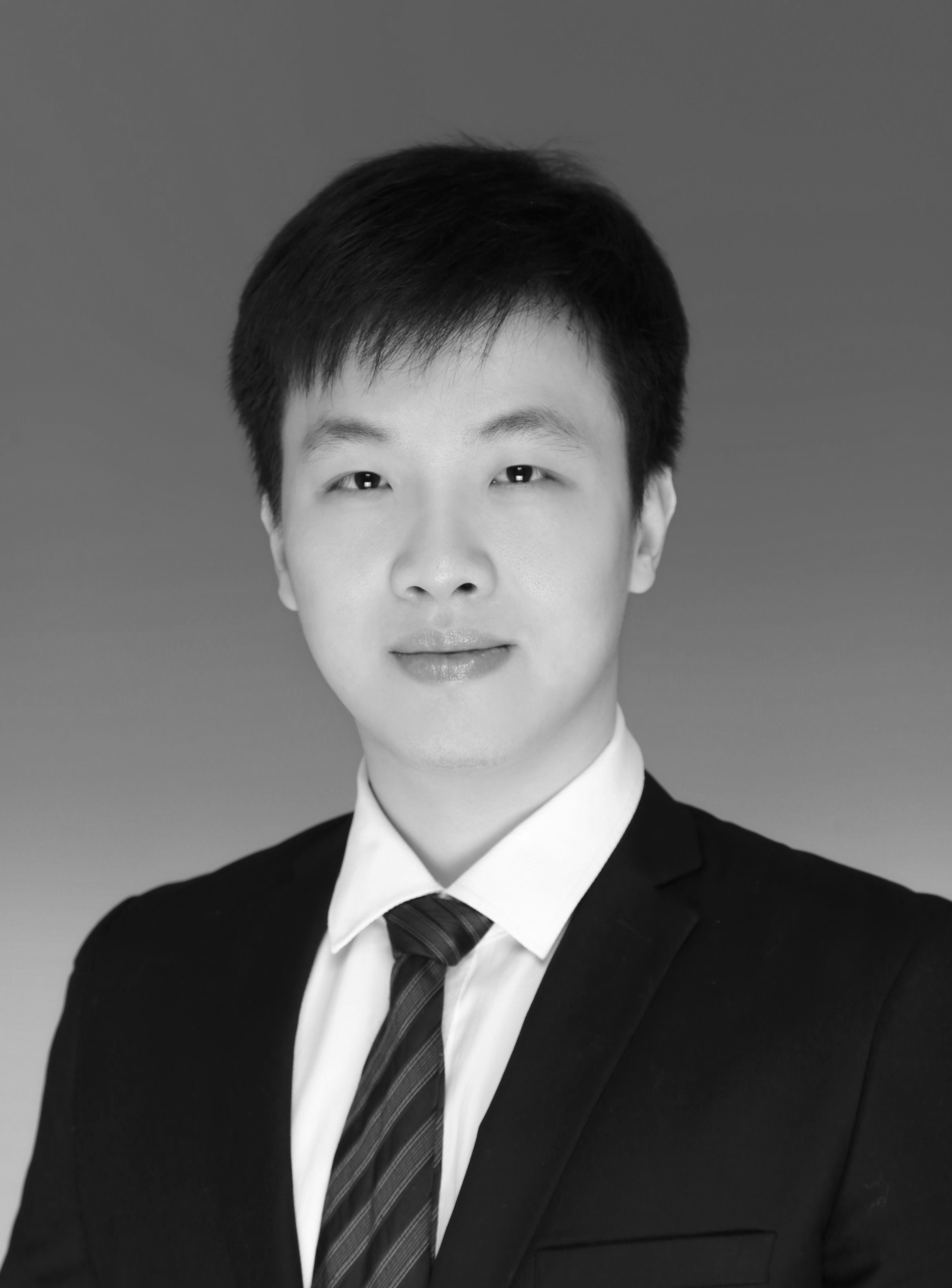}}]{Qi Sun} received his Ph.D. degree in Automotive Engineering from Ecole Centrale de Lille, France, in 2017. He did scientific research and completed his Ph.D. dissertation in CRIStAL Research Center at Ecole Centrale de Lille, France, between 2013 and 2016. He is currently a Postdoctor at the State Key Laboratory of Automotive Safety and Energy and at the Department of Automotive Engineering, Tsinghua University, Beijing, China. His active research interests include intelligent vehicles, automatic driving technology, distributed control and optimal control.
\end{IEEEbiography}

\vskip -2\baselineskip plus -1fil

\begin{IEEEbiography}[{\includegraphics[width=1in,height=1.25in,clip,keepaspectratio]{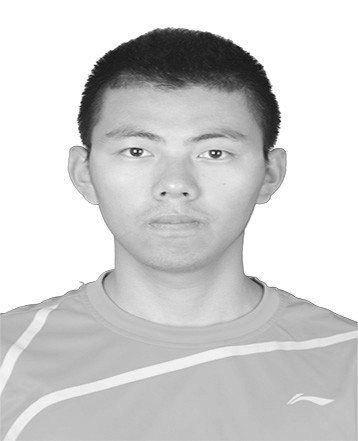}}]{Laiquan Luo}
Laiquan Luo received the M.S. degree from colloge of engineering, China Agricultrual University, Beijing, China, in 2019. He is currently an algorithm engineer at  China Intelligent and Connected Vehicles (Beijing) Research Institute Co.,Ltd. His research interests include computer vision and deep reinforcement learning.
\end{IEEEbiography}

\vskip -2\baselineskip plus -1fil

\begin{IEEEbiography}[{\includegraphics[width=1in,height=1.25in,clip,keepaspectratio]{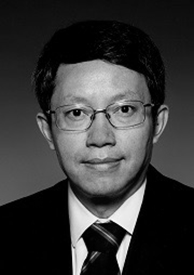}}]{Keqiang Li}
Keqiang Li received his B.E. degree from Tsinghua University, Beijing, China, in 1985 and the M.S. and Ph.D. degree in Chongqing University, Chongqing, China, in 1988 and 1995 respectively. He is a professor in Automobile Engineering at Tsinghua University. He has authored over 90 papers and is co-inventor of 12 patents in China and Japan. His research interest includes vehicle dynamics and control for driver assistance system and hybrid electrical vehicle. He has severed as a Senior Member of the Society of Automotive Engineers of China, and on the editorial boards of International Journal of Vehicle Autonomous Systems. He has received the “Changjiang Scholar Program Professor” and some awards from public agencies and academic institutions of China.
\end{IEEEbiography}

\vskip -2\baselineskip plus -1fil
\end{document}